\pdfoutput=1

\documentclass[11pt]{article}

\usepackage[final]{acl}

\usepackage{times}
\usepackage{latexsym}

\usepackage[T1]{fontenc}

\usepackage[utf8]{inputenc}

\usepackage{microtype}

\usepackage{inconsolata}

\usepackage{graphicx}
\usepackage{booktabs}  
\usepackage{multirow}
\usepackage{colortbl}
\usepackage{xcolor}
\usepackage{tcolorbox}
\usepackage{fontawesome5}
\usepackage{tabularx}

\usepackage{enumitem}

\tcbuselibrary{breakable}

\usepackage{algorithm}
\usepackage{algorithmic}
\usepackage{amsmath}
\usepackage{amssymb}
\usepackage{mathtools}
\usepackage{amsthm}
\theoremstyle{plain}
\newtheorem{theorem}{Theorem}[section]

\theoremstyle{definition}
\newtheorem{definition}[theorem]{Definition}

\newtheorem*{hypothesis}{Hypothesis}

\theoremstyle{remark}

\newcommand{\M}{\mathcal{M}}

\newcolumntype{L}[1]{>{\raggedright\arraybackslash}p{#1}}

\title{Beyond Outlining: Heterogeneous Recursive Planning for Adaptive Long-form Writing with Language Models}

\author{
 \textbf{Ruibin Xiong\textsuperscript{2}\thanks{Equal contribution.}},
 \textbf{Yimeng Chen\textsuperscript{1}\footnotemark[1]\thanks{Corresponding author.}},\\
 \textbf{Dmitrii Khizbullin\textsuperscript{1}},
 \textbf{Mingchen Zhuge\textsuperscript{1}},
 \textbf{Jürgen Schmidhuber\textsuperscript{1,3,4}}
\\
 \textsuperscript{1}Center of Excellence for Generative AI, KAUST
 \textsuperscript{2}Independent Researcher\\
 \textsuperscript{3}The Swiss AI Lab, IDSIA-USI/SUPSI
 \textsuperscript{4}NNAISENSE
\\
  \texttt{ruibinxiong@outlook.com,{yimeng.chen}@kaust.edu.sa}
  \\
    \texttt{\{dmitrii.khizbullin, mingchen.zhuge, juergen.schmidhuber\}@kaust.edu.sa} \\ [6pt]
    \texttt{\href{https://github.com/principia-ai/WriteHERE}{\faGithub \;principia-ai/WriteHERE}}
}

\begin{document}
\maketitle
\begin{abstract}
Long-form writing agents require flexible integration and interaction across information retrieval, reasoning, and composition. Current approaches rely on predefined workflows and rigid thinking patterns to generate outlines before writing, resulting in constrained adaptability during writing. In this paper we propose WriteHERE, a general agent framework that achieves human-like adaptive writing through recursive task decomposition and dynamic integration of three fundamental task types: retrieval, reasoning, and composition. Our methodology features: 1) a planning mechanism that interleaves recursive task decomposition and execution, eliminating artificial restrictions on writing workflow; and 2) integration of task types that facilitates heterogeneous task decomposition. Evaluations on both fiction writing and technical report generation show that our method consistently outperforms state-of-the-art approaches across all automatic evaluation metrics, demonstrating the effectiveness and broad applicability of our proposed framework. We have publicly released our code and prompts to facilitate further research.
\end{abstract}

\section{Introduction}

\begin{figure}[t]
\vskip 0.2in
\begin{center}
\centerline{\includegraphics[width=0.48\textwidth]{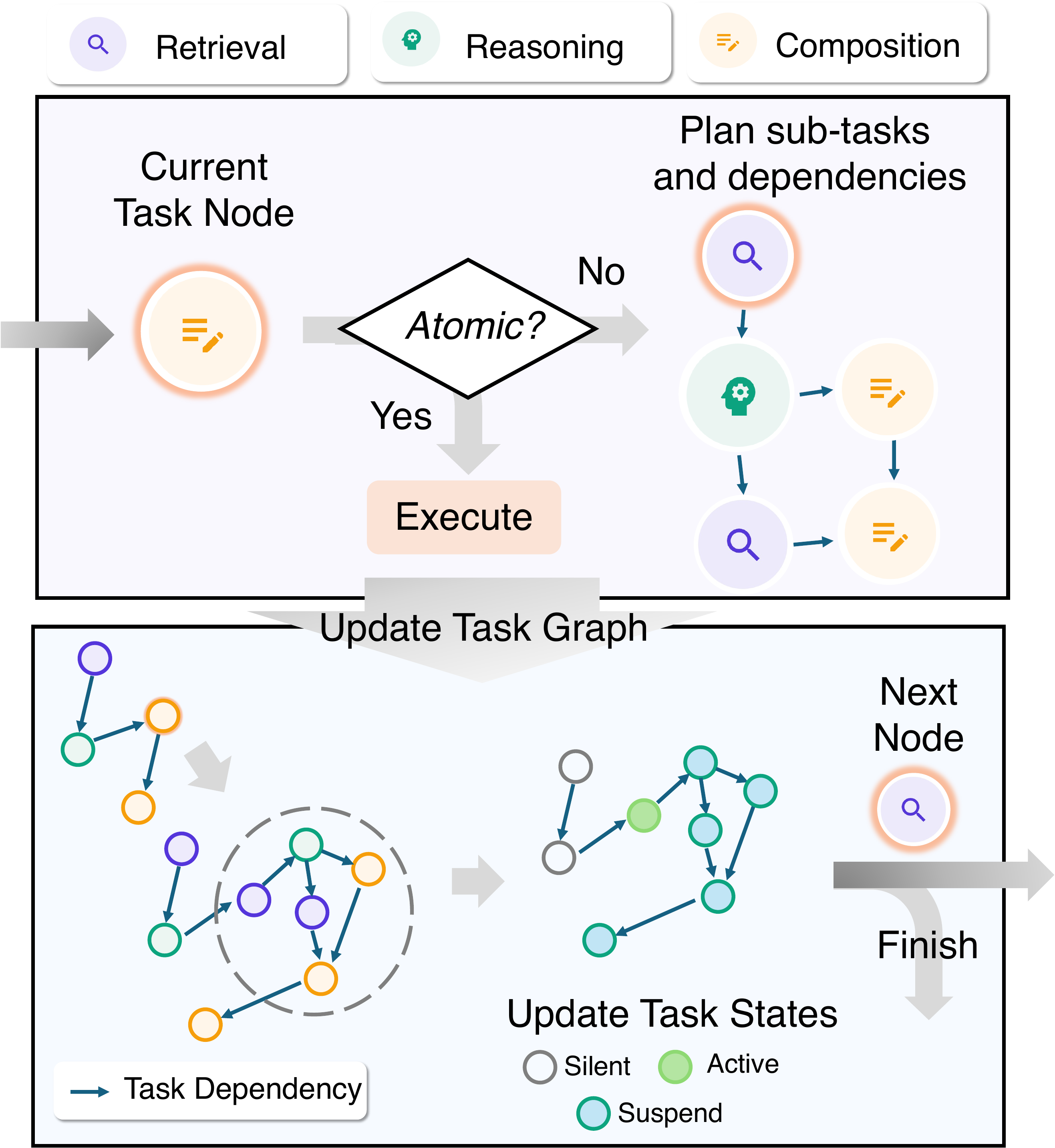}}
\caption{Illustration of the WriteHERE framework for long-form writing. The core of the framework is a heterogeneous recursive planning mechanism that breaks down complex writing goals into primitive subtasks across three cognitive categories. The process is represented as a Directed Acyclic Graph, where a State-based Hierarchical Task Scheduling algorithm manages the adaptive interleaving of task planning and execution.}
\label{fig:overview}
\end{center}
\vskip -0.2in
\end{figure}

Long-form writing plays a crucial role in numerous domains, including narrative generation~\citep{huot2024agents}, academic research~\citep{lu2024ai}, and technical reporting~\citep{shao2024assisting}. Generating coherent, high-quality, and well-structured long-form content presents a significant challenge for Large Language Model (LLM) based writing agents. While LLMs have demonstrated remarkable proficiency in short-form text generation~\citep{yang2022re3,fitria2023artificial}, their ability to sustain consistency, maintain logical coherence, and adapt dynamically across extended passages remains limited~\citep{yang2023doc,bai2024longwriter,huot2024agents}. The complexity of long-form writing arises from the need to manage interdependent ideas, refine arguments progressively, and integrate diverse information sources, all while ensuring stylistic and factual consistency over extended outputs.

Recent advancements in long-form writing have emphasized a pre-writing planning stage to address these challenges~\citep{yang2023doc,huot2024agents,bai2024longwriter,shao2024assisting,jiang2024into}. In the pre-writing phase, an agent first generates a comprehensive outline before proceeding with content generation. For example,~\citet{bai2024longwriter} adopted the plan-and-write paradigm~\citep{yao2019plan} to extend LLM-generated content length by planning the structure and target word count for each paragraph then write paragraphs sequentially. Agent's Room~\citep{huot2024agents} argue that a planning stage is important for narrative generation following the narrative theory and proposed a multi-agent framework to generate the plan and write collaboratively.
STORM~\citep{shao2024assisting} incorporates a multi-agent collaborative outlining stage for retrieval-augmented writing.

However, methods that incorporate a pre-writing stage constrains adaptive reasoning during the writing process. Consider a mystery novelist who discovers an unexpected plot element mid-chapter: they need to retrieve relevant forensic knowledge, reason about plot consistency, and seamlessly integrate new exposition into the narrative flow. Existing structured workflows struggle with such dynamic adjustments since they either have a fixed outline or follow a predefined task sequence. This inflexibility prevents writers from making the necessary modifications when they need to revise their plan and engage in deeper reasoning throughout the writing process.

In this paper, we unify writing and outlining in a general planning framework that enables dynamic adaptation throughout the writing process. We identify three distinct cognitive tasks involved in writing: retrieval, reasoning, and composition, each characterized by unique information flow patterns. Drawing inspiration from Hierarchical Task Network planning (HTN)~\citep{sacerdoti1971structure,georgievski2015}, we formulate long-form writing as a planning problem where the overall writing goal is achieved through the execution of primitive tasks across these three cognitive categories.

Based on the formulation, we propose WriteHERE, a general long-form Writing framework based on HEterogeneous REcursive planning (Figure~\ref{fig:overview}). Leveraging the goal-directed nature of writing tasks, our approach specifies task types during the planning phase and recursively decomposes them into subtasks across the three cognitive categories. This decomposition is recursively applied to subtasks until primitive tasks are reached. The recursive decomposition mechanism enables the system to dynamically adjust planning depth according to the complexity of the writing task and adapt to various requirements. Incorporating task heterogeneity into the planning process facilitates the integration of heterogeneous agents for task execution and type-aware task decomposition.

To enable an adaptive writing process, we interleave task execution with planning. When a primitive task is reached, the system immediately executes it, updates the state of all dependent tasks, and then proceeds to the next task node. To manage this execution and recursive planning procedure, we introduce a State-based Hierarchical Task Scheduling algorithm, where tasks and their dependencies are represented as a Directed Acyclic Graph (DAG). We manage the states of tasks to ensuring a hierarchical and dependency-based execution logic.

While existing methods specified to a fixed scenario, we argue that our method can be generalized across multiple writing tasks. We implement WriteHERE on two distinct long-form writing tasks: technical report generation and narrative generation. Our framework is evaluated on relevant benchmarks, including the \textsc{Tell me a story} dataset for fiction writing and the Wildseed dataset for structured document generation. Experimental results demonstrate that our approach significantly improves content quality and adaptability compared to state-of-the-art baselines.

Our key contributions are as follows.
\begin{itemize}
\item We propose a planning view of the long-form writing problem, casting the process as a combination of heterogeneous tasks that integrates outlining and writing under a single, goal-driven framework.
\item We introduce heterogeneous recursive planning that recursively decomposes tasks into subtasks with specified types, enabling flexible integration of specialized agents and type-aware task decomposition.
\item We develop a State-based Hierarchical Task Scheduling algorithm that efficiently manages adaptive execution and dynamic planning.
\item Experiments on both narrative and report generation show significant improvements of our framework over state-of-the-art baselines.
\end{itemize}

\section{Related Works}
\paragraph{Long-form writing with LLM.} 
Current approaches to long-form generation primarily adopt a multi-stage paradigm, often designed for specific scenarios with limited generalizability. 
Early research by~\citet{yang2022re3,yang2023doc} highlights the significance of comprehensive outlines for story creation. More recently,~\citet{bai2024longwriter} suggested that the output length of LLMs is limited by the SFT data distribution and introduced a Plan-Write framework, which successfully extended GPT-4o's creation to 20,000 words but maintained a static workflow focused solely on length extension.
STORM~\citep{shao2024assisting}, which utilize the autonomous discussion of multi-agents achieved improved factuality through retrieval-augmented outline generation for Wikipedia-like articles, yet its outlines remain fixed once generated. While Co-STORM~\citep{jiang2024into} further incorporated user interaction for outline optimization in report writing, it still lacks the capability to dynamically adjust the writing process. Agent's Room~\citep{huot2024agents} employed multi-agent collaboration but imposed rigid role divisions between planning and writing agents, specifically targeting narrative fiction. Although these approaches successfully address their targeted scenarios, their predetermined workflows not only limit adaptability during writing, but also restrict their applicability across different writing tasks. 

\paragraph{Task decomposition.}
Neural networks for task decomposition can facilitate long-term sequential planning and decision-making by discovering sub-problems and exploiting sub-solutions~\cite{SchmidhuberWahnsiedler:92sab}.
Recent research demonstrates that incorporating task decomposition during LLM inference improves performance on language tasks.
~\citet{wei2022chain} showed that explicit chain-of-thought task decomposition during inference significantly enhances the capabilities of LLMs. Approaches like least-to-most prompting~\citep{zhou2022least} and ReAct~\citep{yao2022react} explicitly interleave task execution and decomposition, while ReasonFlux~\citep{yang2025reasonflux} proposed a template-based method for generating reasoning trajectories. For long-form writing, flat planning methods face challenges, as the complex hierarchical dependencies within linear context history can become unwieldy and lead to a loss of coherence.
Other works have explored hierarchical decomposition approaches. For example,~\citet{khot2023decomposed} designed a modular planner-executor system with distinct few-shot prompts that can recursively decompose tasks into smaller problems of the same form. ADaPT~\citep{prasad2023adapt} introduced on-demand recursive decomposition, yet did not address the integration of fundamentally different types of operations such as retrieval and reasoning. These existing methods primarily focused on the reasoning tasks. In this work, we propose a heterogeneous recursive framework that effectively handles long-form writing tasks with distinct operational characteristics.
Our goal-decomposition approach is also distinct from and complementary to path exploration methods like ToT~\citep{yao2023tree}, CoR~\citep{wang2025cor}, which are focused on explore multiple parallel reasoning paths to optimize a single step.

\section{Formulation}
In this section, we formulate the fundamental components of a long-form writing agent system, focusing on three heterogeneous task types essential for writing: retrieval (information gathering), reasoning (content planning), and composition (text generation). We further formalize the writing planning problem with a conceptual framework inspired by the hierarchical task network planning.

\subsection{Writing Agent System}
We first introduce the notion of the writing agent system.

\begin{definition}[Writing Agent System]
\label{def:agent-system}
A \emph{writing agent system} is a tuple
\[
  \Sigma_{\mathcal{A}} \;=\; (\mathcal{A}, \mathcal{M}, D, W),
\]
where \(\mathcal{A}\) is the \emph{agent kernel} responsible for processing writing instructions, solving writing tasks, and selecting actions. \(\mathcal{M}\) is the \emph{internal memory} maintaining writing-related information like outlines, drafted content, and retrieved references. $D$ is the \emph{database} (e.g., search engine, reference documents) and $W$ is the writing \emph{workspace}. 
\end{definition}

\begin{figure}[t]
\vskip 0.2in
\begin{center}
\centerline{\includegraphics[width=0.5\textwidth]{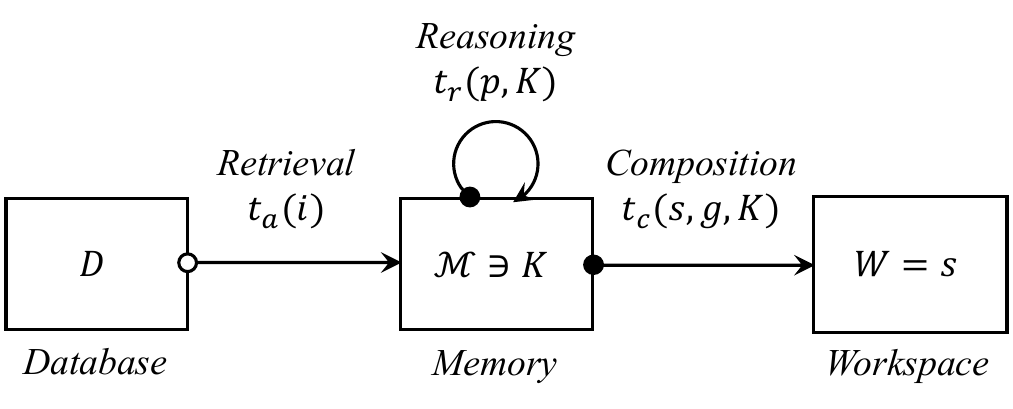}}
\caption{The abstract flow of tasks. The arrow indicates the information flow of a task: the system state at the arrowhead is modified by the labeled task, while the hollow circle end signifies that the associated system state remains unchanged.}
\label{fig:tasks}
\end{center}
\vskip -0.2in
\end{figure}

\subsection{Task Types}
\label{subsec:def_types}
The writing process naturally involves three types of heterogeneous cognitive tasks: retrieval for information gathering, reasoning for content planning, and composition for content generation. This categorization aligns with cognitive models of agents~\citep{sumers2024cognitive} and reflects the distinct operational patterns in writing tasks.

\begin{definition}[Retrieval Task]
Let $i$ be the information needs during writing (e.g., factual queries, reference searches). A \emph{retrieval task} $t_a(i)$ for aims to acquire relevant information from the environment and update it into the agent's memory $\mathcal{M}$. 
\end{definition}

\begin{definition}[Reasoning Task]
Let $p$ represent a writing-related problem requiring logical inference (e.g., outline planning, content organization). A \emph{reasoning task} $t_r(p, K)$ aims to derive new knowledge or make decisions based on available information $K$ in agent's internal memory $\M$. 
\end{definition}

\begin{definition}[Composition Task]
Let $g$ represent the text generation objective specifying target states of the written content. A \emph{composition task} $t_c(s, g, K)$ aims to generate text that meets specified requirements (e.g., style, length, structure) through a sequence of writing actions, given current workspace state $s$ and knowledge $K \in \M$. 
\end{definition}

We illustrate the abstract flow of the three tasks in Figure~\ref{fig:tasks}. Retrieval Task functions as context-independent operations that enhance working memory without modifying the workspace; Reasoning Task performs memory-to-memory transformation contingent upon satisfaction of logical preconditions; and Composition Task executes workspace-altering operations and then updates related information to the memory.

\subsection{Planning for Writing}
\label{subsec:planning_problem}
Planning for writing is based on the assumption that the writing process as complex tasks composed by simpler, executable subtasks. This perspective follows HTN planning, where the objective is not to achieve a set of goals but instead to perform some set of primitive tasks. 

In the context of writing, primitive tasks are the basic actions that can be executed directly by the agent. Breaking down complex tasks into these primitives improves accuracy~\citep{chen2024unlocking} and allows flexible action interleaving. By assuming a theoretical set $T_p$ of primitive tasks (without explicitly specifying its composition), we formulate the writing planning problem as follows.

\begin{definition}[Writing Planning Problem]
\label{def:writing-planning}
A \emph{writing planning problem} is a tuple 
\[
  \langle\, t_c(g, s_0, K_0),\; T_p \,\rangle,
\]
where $t_c(g, s_0, K_0)$ is the top-level composition task, with a writing goal $g$, the initial state of the writing workspace \(s_0\), and the initial content of the agent's memory $K_0$. \(T_p\) is the set of executable primitive 
retrieval, reasoning and composition tasks. A solution \(\pi = \langle t_1, t_2, \dots, t_k \rangle\) to this planning problem is a sequence of primitive tasks that achieves the writing objective while maintaining coherence and satisfying constraints.
\end{definition}

\section{Heterogeneous Recursive Planning}
Based on the formulation of the writing task planning problem, we propose a heterogeneous recursive planning method (HRP) inspired by the HTN planning and the heterogeneity of the three cognitive tasks. In this section, we introduce the key components of our approach.

\subsection{Recursive Planning}
The classical HTN planning paradigm solves problems through hierarchical decomposition until reaching primitive executable operations. Following our formulation of the writing planning problem, we adopt a recursive planning strategy, in alignment with classical HTN approaches.

The core of this planning process is task decomposition: each task is broken down into subtasks, and the same decomposition logic is recursively applied to those subtasks. Unlike traditional as-needed decomposition methods that rely on execution failure to stop further planning, our approach introduces a different termination criterion. We only continue planning if certain subtask types necessitate further decomposition, ensuring that the final operations are always executable without redundant decomposition.

\subsection{Typed Task Integration}
\label{subsec:typed}
Building upon our formal characterization of cognitive task types in Section~\ref{subsec:def_types}, we extend the recursive planning framework with type-aware decomposition mechanisms.

Our integration addresses the cognitive heterogeneity inherent in writing processes. While complex tasks may involve blended operations, their decomposition should respect the dominant cognitive type based on primary objectives. We formalize this as:
\begin{hypothesis}[Type Specification in Decomposition]
During hierarchical decomposition of writing tasks, all generated subtasks can be specified as exactly one cognitive type.
\end{hypothesis}
This hypothesis suggests that the writing planning problem can be decomposed into sub-planning problems of three distinct task types. For example, assume task $t_c(g,s_0, K_0)$ can be decomposed into a sequential combination of subtasks $t_a(i)$, $t_r(p, K')$, and $t_c(g, s_0, K'')$, where $K'$ and $K''$ denote the modified knowledge in $\M$ after executing the preceding tasks. The solution of $\langle t_c(g,s_0, K_0), T_p \rangle$ is then the combination of solutions of planning problems $\langle t_a(i), T_p\rangle$, $\langle t_r(p, K'), T_p\rangle$, and $\langle t_c(g, s_0, K''), T_p\rangle$. These solutions must satisfy their corresponding executability conditions and goal achievement criteria. For instance, subtasks of composition may include retrieval or reasoning tasks to modify the internal memory. They must have a composition-type subtask to reach the goal.

Motivated by the above analysis, we integrate task types into the planning procedure. Our method features the following key design elements:
\begin{itemize}
\item \textbf{Dynamic type annotation}: Each subtask generated in a planning step is assigned a specific type. It facilitates the function call of heterogeneous agents, for example, a search agent to conduct a retrieval task.
\item \textbf{Type-aware decomposition}: This provides targeted guidance for potential subtask breakdowns based on the type of the current task.
\end{itemize}

\begin{algorithm}[t]
\caption{WriteHERE framework}
\label{alg:DHTS}
\begin{algorithmic}[1]
\REQUIRE Memory $\mathcal{M} = (G, W)$: Task Graph $G=(V,E)$ with root $V_{\text{init}} = \{v_{\text{root}}\}$; Workspace $W$; Initial state $S(v_{\text{root}}) \gets \textsc{Active}$
\ENSURE $S(v) =\textsc{Silent}$, $\forall v \in V$
\WHILE{$\exists v \in V \mid S(v) \neq \textsc{Silent}$}
    \STATE Select $v^* \gets \arg\min_{v \in V} \{\text{BFS-depth}(v) \mid S(v)=\textsc{Active}\}$
    \STATE Get knowledge $K \gets$ \textsc{GetInfo}($\mathcal{M}, v^*$)
    $v^* \gets$ \texttt{Update}($v^*, K$)
    \IF{\texttt{IsAtomic}($v^*, K$)}
        \STATE $M \gets$ \texttt{Execute}($v^*, K$) // Differs depending on task type
        \STATE $S(v^*) \gets$ \textsc{silent}
    \ELSE
        \STATE $\{v_1,\dots,v_k\} \gets $ \texttt{TypedPlan}($v^*, K$)
        \STATE \textsc{AddChildren}($G, \{v_1,...,v_k\}, v^*$)
        \STATE $S(v^*) \gets$ \textsc{Suspended}
    \ENDIF
    \STATE Update $S(v)$ for all $v$ in $V$ to \{\textsc{Silent}, \textsc{Suspended} or \textsc{Active}\}
\ENDWHILE
\end{algorithmic}
\end{algorithm}

\section{WriteHERE Framework}
We propose WriteHERE, an adaptive writing framework that integrates HRP with state-based hierarchical task scheduling, implemented using structural memory and graph-based context control. We summarize its core logic in Algorithm~\ref{alg:DHTS} and introduce the key concepts below. A detailed walkthrough with a specific example is provided in Appendix~\ref{apsec:walk}.

\paragraph{Task graph.} Tasks and their dependencies are modeled as a directed acyclic graph $G=(V,E)$.
Each node is denoted with the type, goal, dependencies information and execution result of it. The graph $G$ starts with a single root node with $g_{root}$ describing the user input request and $t_{root}$ defined as composition. $G$ is dynamically expanded and updated throughout the process.

\paragraph{State-based hierarchical task scheduling.}
Our approach interleaves task execution with planning, enabling adaptive planning that responds to action outcomes through a hierarchical task scheduling algorithm.
The algorithm manages dynamic task decomposition through assigning one of the three states to each task node $v$, denoted as $S(v)$: \textsc{Active}, \textsc{Suspended}, or \textsc{Silent}. A task is \textsc{Suspended} while its prerequisites are incomplete or after it has been decomposed into subtasks. It becomes \textsc{Active} only when all prerequisites are met, marking it ready for processing. Upon completion, a task transitions to the \textsc{Silent} state.
Starting from the root, the algorithm iteratively selects \textsc{Active} task nearest to the root with BFS-based topological sorting. The selected task is either executed directly (if primitive) or decomposed into subtasks which are then integrated into the graph. This process continues until all tasks reach the \textsc{Silent} state, ensuring the systematic completion of the entire task hierarchy.

\paragraph{Memory and context control.} The memory $\mathcal{M}$ of our agent system consists of task graph $G$ and the workspace $W$. This memory does not serve as the complete context for planning or subtask execution; instead, relevant knowledge is retrieved through a context control module.
As introduced in Section~\ref{subsec:typed}, the knowledge context of a decomposed subtask is determined by the knowledge context of its parent task and the execution results of its preceding tasks. Our context control strategy adheres to this principle.
For each task node, the framework constructs task-specific knowledge comprising the current workspace state and relevant task graph information, including node information from parent nodes up to a specified depth and precedent nodes on which it depends. Additionally, the planning modules (\texttt{IsAtomic} and \texttt{TypedPlan}) receives global structural information about $G$, including the goals, types, and dependencies of all nodes. We abstract this logic as \textsc{GetInfo}$(\mathcal{M}, v)$ in Algorithm~\ref{alg:DHTS}.

\paragraph{LLM operations.} The framework prompting LLMs for the following core operations: updates the task goals, determines the primitivity of the task, execute the primitive task, and generate the typed plan. Specifically, the \texttt{Update} module refines the goal of the selected task node based on the related knowledge. The \texttt{IsAtomic} module then employs an LLM to determine if a task is atomic (i.e. primitive, directly executable) or complex (requiring decomposition). If a task is complex, the \texttt{TypedPlan} module decomposes the goal into a structured list of subtasks. To ensure validity, this process employ structured prompting to constrain the LLM's output format and apply programmatic validation rules to detect and correct dependency errors, guaranteeing robust execution.
The \texttt{Execute} module invokes specialized executors for different primitive task types. Specifically, the composition executor generates text segments, while the reasoning executor produces structured analyses or decisions. The retrieval executor returns a summary of the retrieved information.

\section{Experiments}
We evaluate our approach through experiments on two challenging long-form writing tasks: narrative generation and report generation. Our investigation addresses three key aspects: (1) the comparative performance of our method against state-of-the-art baselines, (2) the impact of the recursive planning and task-type module, and (3) the generalization capability across diverse task domains. 

\begin{table*}[ht]
\centering
\resizebox{0.92\textwidth}{!}{
\begin{tabular}{llccccc}
\toprule
\multirow{2}{*}{\textbf{Backbones}} & \multirow{2}{*}{\textbf{Methods}} & \multicolumn{4}{c}{\textbf{Dimensions}} & \\ \cmidrule(lr){3-7}
& & Plot & Creativity & Development & Language Use & Overall \\
\midrule
\multirow{5}{*}{GPT-4o} 
& E2E         & 0.337  & 0.218  & 0.288  & 0.202  & 0.270  \\
& Agent's Room & 1.035  & 0.712  & 0.948  & 0.680  & 0.869  \\
& WriteHERE  & \textbf{1.470}  & \textbf{2.005}  & \textbf{1.967}  & \textbf{2.233}  & \textbf{2.143}  \\
& \cellcolor{gray!10}\;w/o Recursive    & \cellcolor{gray!10}1.307   & \cellcolor{gray!10}1.327         & \cellcolor{gray!10}1.041        & \cellcolor{gray!10}1.192           & \cellcolor{gray!10}1.100      \\
& \cellcolor{gray!10}\;w/o Type        & \cellcolor{gray!10}0.852 & \cellcolor{gray!10}0.733  & \cellcolor{gray!10}0.756       & \cellcolor{gray!10}0.693        & \cellcolor{gray!10}0.717     \\
\midrule
\multirow{5}{*}{Claude-3.5-Sonnet}
& E2E         &    0.036  &       0.016    &      0.032       &      0.017        &   0.025     \\
& Agent's Room &   1.029   &      0.480     &      0.778       &     0.484         &    0.694     \\
& WriteHERE  & \textbf{2.016}   & \textbf{2.634}      & \textbf{2.959}       & \textbf{2.264}         & \textbf{2.852}     \\
& \cellcolor{gray!10}\;w/o Recursive    & \cellcolor{gray!10}1.145   & \cellcolor{gray!10}1.396        & \cellcolor{gray!10}0.707         & \cellcolor{gray!10}1.517      & \cellcolor{gray!10}0.918      \\
& \cellcolor{gray!10}\;w/o Type        & \cellcolor{gray!10}0.774   & \cellcolor{gray!10}0.475       & \cellcolor{gray!10}0.525        & \cellcolor{gray!10}0.518         & \cellcolor{gray!10}0.512  \\
\bottomrule
\end{tabular}
}
\caption{Quantitative strength scores of methods on the \textsc{Tell me a story} dataset. The scores are derived from pairwise comparisons of all generated stories, with the final relative strength calculated using the Davidson model. This score is non-linear; improvements at the higher end of the scale are progressively more challenging. Ablations of our method are highlighted in grey. The highest value in each column is in bold.}
\label{tab:main_story}
\end{table*}

\begin{figure*}[ht]
    \centering
    \includegraphics[width=0.95\textwidth]{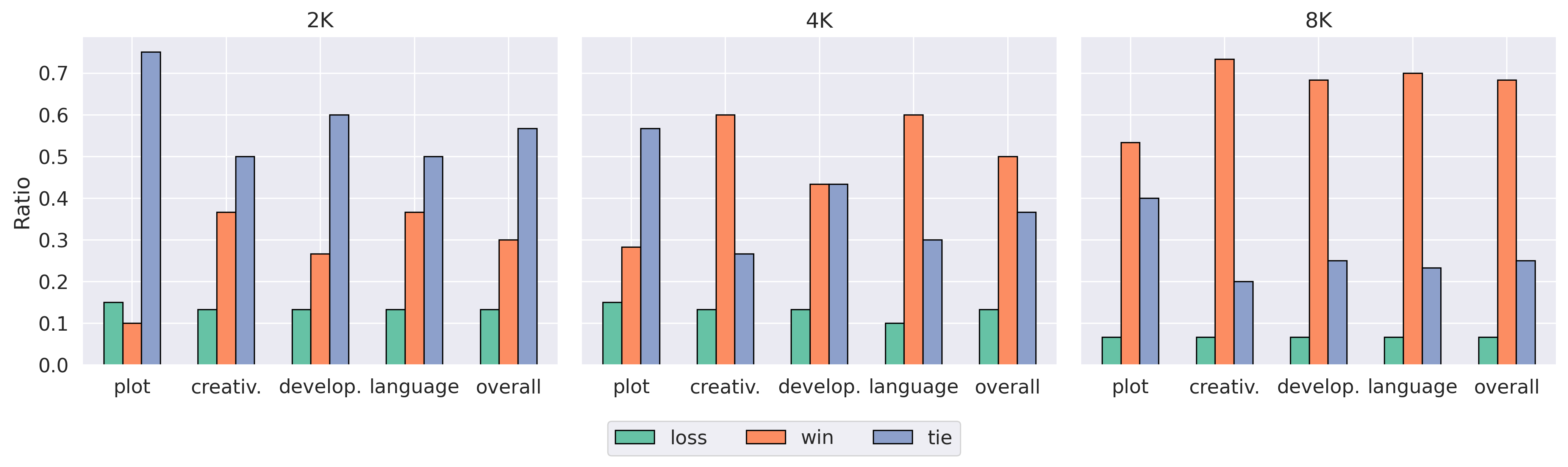} 
    \caption{The evaluation results of WriteHERE v.s. Agent's Room at different generation lengths.}
    \label{fig:length}
\end{figure*}

\subsection{Narrative Generation}
Narrative generation involves complex reasoning and composition tasks. We use the \textsc{Tell me a story} fiction writing dataset proposed in the paper of Agent's Room~\citep{huot2024agents}.

\paragraph{Datasets.} \textsc{Tell me a story} offers a collection of complex, well-structured narratives paired with detailed narrative generation prompts. The dataset consists of 230 samples, with each prompt averaging 113 tokens and corresponding narrative responses averaging 1,498 tokens.

\paragraph{Baselines.} We implement two primary baselines: (1) End-to-End (E2E): where we directly provide the story prompt to the base LLM without any additional guidance or planning steps; and (2) Agents' Room~\citep{huot2024agents}: a collaborative writing framework with multiple agents that decomposes the story generation process into planning and writing phases. In the planning phase, specialized agents outline key story elements including plot structure, character development, and setting details. Writing agents then generate the full narrative following this structured plan. 

\paragraph{Evaluation metrics.} We adopt the LLM-based evaluator for story assessment proposed by \citet{huot2024agents}, which demonstrates strong correlation with human judgments (Spearman's rank correlation $\rho = 0.62, p < 0.01$). For each story pair, the evaluator determines which is superior or equivalent across these dimensions and overall, producing win-tie-loss judgments. To convert these pairwise comparisons into quantitative scores, we employ the Davidson model~\citep{davidson1970extending}, which effectively handles cases with ties. Following the practice of~\citet{huot2024agents}, we implement the evaluator using Gemini (2.0-Flash) as the base LLM.  To mitigate position bias, we conduct 7 evaluations in each ordering (14 total trials) and determine the final outcome through majority voting.

\begin{table*}[ht]
\centering
\resizebox{0.78\textwidth}{!}{
\begin{tabular}{llcccc}
\toprule
\multirow{2}{*}{\textbf{Backbones}} & \multirow{2}{*}{\textbf{Methods}} & \multicolumn{4}{c}{\textbf{Report Quality}} \\
\cmidrule(lr){3-6}
& & Relevance & Breadth & Depth & Novelty \\
\midrule
\multirow{4}{*}{GPT-4o} 
& STORM    &  4.76  & 4.58 &  4.30 &  4.32   \\
& Co-STORM & 4.36 & 4.22 & 4.02 & 4.17 \\
& WriteHERE & \textbf{4.93} & \textbf{4.86} & \textbf{4.79} & \textbf{4.51} \\
& \cellcolor{gray!10}\;w/o HRP & \cellcolor{gray!10}4.83 & \cellcolor{gray!10}4.18 & \cellcolor{gray!10}3.74 & \cellcolor{gray!10}4.17 \\
\midrule
\multirow{4}{*}{Claude-3.5-Sonnet}
& STORM   & 4.66  &  4.63 & 4.40 &  4.41   \\
& Co-STORM & 3.87 & 3.56 & 3.46 & 3.82 \\
& WriteHERE   & \textbf{4.96} & \textbf{4.92} & \textbf{4.93} & \textbf{4.82} \\
& \cellcolor{gray!10}\;w/o HRP & \cellcolor{gray!10}4.84 & \cellcolor{gray!10}4.51 & \cellcolor{gray!10}4.24 & \cellcolor{gray!10}4.46 \\
\midrule
\multirow{2}{*}{DeepSeek-R1}
& WriteHERE & \textbf{4.97} & \textbf{4.94} &  \textbf{4.95} & \textbf{4.88} \\
& \cellcolor{gray!10}\; w/o HRP & \cellcolor{gray!10}4.94 & \cellcolor{gray!10}4.81 & \cellcolor{gray!10}4.83 & \cellcolor{gray!10}4.80 \\
\midrule
Commercial & PPL-Deep Research & 4.93 & 4.73 & 4.75 & 4.45 \\
\bottomrule
\end{tabular}
}
\caption{Comparison of method performance on WildSeek, evaluated by \texttt{o1-preview}. The scores represent absolute grades on a 1-5 scale based on a detailed rubric. Our method and its ablations are highlighted with a grey background.}
\label{tab:rubric_comparison}
\end{table*}

\paragraph{Configurations.} For Agent's Room baseline, we implement the plan+write version according to the paper, which includes 4 planning agents (conflict, character, setting, plot) and 5 writing agents (exposition, rising action, climax, falling action, resolution). We use a length estimator along with the writing agents to enable the length control. For our method, two task types are included: reasoning (Design) and composition (Writing). We implement a Design agent and a Writing agent as the primitive task executors.

\subsubsection{Results}
As shown in Table~\ref{tab:main_story}, Agent's Room significantly outperforms the E2E baseline, aligning with results reported in their original paper. Our proposed method demonstrates superior performance across all five key evaluation metrics compared to baseline approaches. This consistent improvement holds across two different backbone LLMs, validating the robustness of our approach across base models.

\paragraph{Ablation study.} To analyze the contributions of individual components, we conducted an ablation study with two key variations:
1) Non-recursive generation (``w/o Recursive"): This variant removes the recursive decomposition process, instead generating the entire plan in a single step similar to baseline methods. 2) Task-type removal (``w/o Type"): This variant omits explicit task-type information during decomposition. While still employing recursive breakdown, the model no longer utilizes type-specific decomposition logic.

\paragraph{Extended lengths.} We also evaluated how different methods scale with increasing generation length. From our dataset, we selected 60 samples that an LLM identified as suitable for generating texts over 8,000 words. We then conducted experiments by prompting models to generate articles of three different lengths: 2K, 4K, and 8K words, operating under the assumption that task complexity increases with required text length. Figure~\ref{fig:length} presents pairwise comparisons of the overall metric between our method and Agents Room with GPT-4o as the base LLM. We excluded the E2E baseline from this comparison as it is unable to generate texts of 4K or 8K words. For 2,000-word stories, our method and Agents Room performed comparably on more than 50\% of samples. However, our method demonstrates increasingly significant advantages over the baseline as task length increases, highlighting its effectiveness in handling more complex long-form content generation.

\subsection{Report Generation}
Compared with story generation, report generation task further need the integration of complex retrieval tasks with reasoning and composition. We employed a hybrid evaluation strategy to balance rigor, scale, and alignment with existing benchmarks. Specifically, we used LLM-based evaluation to enable large-scale pairwise comparisons and human evaluation for the most challenging and complex reports over 10,000 words.

\paragraph{Datasets.} We use the WildSeek dataset proposed by~\citep{jiang2024into}. WildSeek offers a collection of real-world information-seeking tasks paired with user goals for evaluating complex information retrieval capabilities. The dataset consists of 100 samples across 24 domains, collected from users of the STORM web application. Each data point comprises a Topic-Intent sentence pair. 

\paragraph{Baselines.} We compare our method with STORM~\citep{shao2024assisting} and Co-STORM~\citep{jiang2024into}. STORM is a writing system that uses perspective-guided question asking from retrieval and constructs Wikipedia-like articles through generating outlines and section-by-section writing. Co-STORM extends STORM by introducing a user-participated roundtable discussion to enhance the diversity of retrieved information and improve coverage of unknown unknowns. Both baseline methods rely on retrieval-augmented generation and use similar outline-driven approaches for long-form text generation.

\paragraph{Evaluation metrics.} We utilize the evaluation framework established by Co-STORM, which examines the final report across four dimensions: Relevance, Broad Coverage (Breadth), Depth, and Novelty. A LLM-based evaluator assesses each dimension on a 5-point scale, with the original Topic and Intent provided. We employ the latest \texttt{OpenAI o1-preview} as our primary evaluator model.

\paragraph{Configurations.} We use Bing Search API for retrieval. We use the latest official implementation of STORM\footnote{https://github.com/stanford-oval/storm} with their default configurations. For Co-STORM, we follow the official implementation with its user-simulator. We design a search agent, an analyzing agent, and a writing agent as the primitive task executors for retrieval, reasoning and composition respectively. For the search agent, we implement a multi-agent framework comprising a retrieval agent, a reranking agent, and a summarization agent. See Appendix~\ref{apsec:details} for more details.

\subsubsection{Results}
Our primary experiment on the WildSeek dataset is presented in Table~\ref{tab:rubric_comparison}. The results demonstrate that our method consistently outperforms the current state-of-the-art approaches across four distinct automatic evaluation metrics. This further validates the effectiveness and generalizability of our approach. We observe a significant improvement in writing depth with our method. Additionally, our approach consistently outperforms existing methods in terms of relevance, engagement, and breadth of the generated content.

\paragraph{Ablation study.} To further validate the effectiveness of our approach, we implemented an ablation version, where we retained the same search agent setup but removed the recursive planning strategy (denoted as ``w/o HRP" in Table~\ref{tab:rubric_comparison}). This modification required the planner to generate subtasks as a linear workflow all at once rather than in a hierarchical manner. By isolating this variable, we could quantify the performance gains specifically attributable to recursive planning. We observe a significant drop in depth metrics in the ablation version, demonstrating the benefits of HRP. Additionally, removing recursive planning results in a notable decline in novelty and breadth, further highlighting its contribution to the generation quality. 

\paragraph{Reasoning model compatibility.} We further experimented using the reasoning model DeepSeek-R1~\citep{deepseekr1} as the base LLM. Results demonstrate that our approach maintains significant performance advantages. Particularly notable improvements were observed in reasoning depth and breadth metrics. This demonstrates our method's consistent ability to enhance reasoning capabilities. Our analysis included Perplexity's Deep Research\footnote{https://www.perplexity.ai/hub/blog/introducing-perplexity-deep-research} (Feb. 2025), a commercial reasoning model based agent, tested on the same dataset. The results demonstrate that our methodology, when implemented with either Claude or DeepSeek-R1 as the base model, delivers significantly superior performance across all measured metrics compared to this commercial alternative.

\subsubsection{Long Reports and Human Evaluation}
To assess our framework's ability in generating extended long-form reports (over 10,000 words), we conducted a dedicated human evaluation study, detailed in Appendix~\ref{apsec:human}. 
\paragraph{Dataset.} Existing datasets like WildSeek provide prompts that are too concise and lack necessary details to specify requirements for complex, long-form reports. To tackle that, we created a new benchmark dataset, LongReport, specifically designed with 12 complex prompts intended to elicit comprehensive reports. Our topic selection prioritizes time-sensitive subjects that require the model to access current knowledge, with topics systematically categorized based on varying assessment emphases. 
\paragraph{Evaluation metrics and baseline.} We adopted the four dimension as on WildSeek with one additional dimension \emph{Clarity, Cohesion, and Language} to assess organization and language use. We recruited five volunteer annotators with qualified technical backgrounds to compare reports generated by WriteHERE against a state-of-the-art commercial baseline, Gemini Deep-Research (2.5 Pro)\footnote{https://gemini.google/overview/deep-research/}. Each annotator provided absolute scores on a 1-5 scale for all five dimensions and indicated their overall preference.

\paragraph{Results.} The results demonstrate that our method exhibits performance comparable to Gemini, with a slight advantage reflected in a 7:5 vote score across 12 topics, which further validates the capability of our approach for long-form writing.

\section{Conclusion}
In this work, we introduce a general framework for long-form writing agents built on heterogeneous recursive planning. Our approach is based on an analysis of three distinct types of tasks in the writing process and formulation of the writing planning problem. 
We highlight the heterogeneity of writing planning, not only in the final generated plan but also in the sub-planning problems that emerge during hierarchical decomposition. To address this, we incorporate type specification into the recursive planning process. Additionally, we employ a state-based task scheduling algorithm for adaptive task execution.
Experiments across narrative and report generation demonstrate significant quality improvements over state-of-the-art baselines, while ablations confirm the critical contributions of both recursive planning and task-type awareness.

\section*{Limitations}
\paragraph{Computational efficiency.} The recursive decomposition process introduces additional computational overhead compared to end-to-end approaches. Future work could explore optimization techniques. Another potential avenue for improving efficiency is the use of heterogeneous agents, where models are assigned to different tasks based on their complexity. Instead of applying a single large model to all recursive decomposition and execution steps, specialized models could be leveraged for simpler subtasks, reserving larger models for more complex reasoning. Furthermore, a reasoning budget could be implemented to explicitly control resource allocation, for instance, by limiting the maximum recursion depth of the task graph or the total number of generated subtasks.

\paragraph{Human-in-the-loop integration.} While our approach automates task decomposition and execution, integrating human feedback during the planning and writing stages could further improve adaptability and quality. Future research could explore interactive refinement mechanisms where users can edit the task graph during planning or guide the generation by feedback.

\paragraph{Process diagnostics.} Our framework would benefit from a process debugging suite. However, its design provides a strong foundation for failure analysis, as the explicit and structured task graph enables the precise tracing of any failure back to its source node or sequence of nodes. Future work could build directly on this traceability by implementing self-correcting methods that use diagnostic feedback to enhance workflow efficiency.

\section*{Acknowledgment}
The research reported in this publication was supported by funding from King Abdullah University of Science and Technology (KAUST) - Center of Excellence for Generative AI, under award number 5940.
The authors are grateful to Zhengying Liu and Piotr Piękos for their insightful discussions and suggestions, and to Dandan Guo, Liangyu Wang, Zheng Zeng, and Han Qian for their valuable assistance with this project. We also extend our gratitude to the anonymous reviewers for their constructive comments, which significantly improved the manuscript.

\bibliography{paper}

\clearpage
\appendix
\onecolumn

\section{Extended Related Works}
\paragraph{Long-form writing with LLM.} 
Current approaches to long-form generation primarily adopt a multi-stage paradigm, often designed for specific scenarios with limited generalizability. 
Early research by~\citet{yang2022re3,yang2023doc} highlights the significance of comprehensive outlines for story creation. More recently,~\citet{bai2024longwriter} suggested that the output length of LLMs is limited by the SFT data distribution and introduced a Plan-Write framework, which successfully extended GPT-4o's creation to 20,000 words but maintained a static workflow focused solely on length extension.
STORM~\citep{shao2024assisting}, which utilize the autonomous discussion of multi-agents achieved improved factuality through retrieval-augmented outline generation for Wikipedia-like articles, yet its outlines remain fixed once generated. While Co-STORM~\citep{jiang2024into} further incorporated user interaction for outline optimization in report writing, it still lacks the capability to dynamically adjust the writing process. Agent's Room~\citep{huot2024agents} employed multi-agent collaboration but imposed rigid role divisions between planning and writing agents, specifically targeting narrative fiction. Although these approaches successfully address their targeted scenarios, their predetermined workflows not only limit adaptability to emergent needs during writing (e.g., contextual conflicts), but also restrict their applicability across different writing tasks. 

\paragraph{Task decomposition.} 
Task decomposition has been a fundamental approach in planning since the introduction of Hierarchical Task Network (HTN) planning~\citep{sacerdoti1971structure}. An HTN planner recursively decomposes nonprimitive tasks into smaller subtasks until reaching primitive tasks that can be performed directly using planning operators. This method has proven particularly effective in real-world applications by explicitly encoding task hierarchies and constraints~\citep{ghallab2004automated,georgievski2015}, and is shown to be more expressive than classical planning~\citep{erol1994htn,ghallab2004automated}. Early systems like NOAH~\citep{sacerdoti1975nonlinear} and Nonlin~\citep{tate1977generating} established the foundations for task decomposition and constraint management, influencing later planners such as SIPE~\citep{wilkins1990can} and O-Plan~\citep{currie1991plan,tate1994plan2}. The SHOP family~\citep{nau1999shop,nau2003shop2} demonstrated impressive performance in real-world tasks through domain-specific decomposition methods, though its heavy reliance on domain knowledge has raised concerns about generalizability~\citep{nau2007current}.

Neural networks for task decomposition can facilitate long-term sequential planning and decision-making by discovering sub-problems and exploiting sub-solutions~\cite{SchmidhuberWahnsiedler:92sab}. Sec. 5.3 of~\citep{schmidhuber2015learning} describes an adaptive ``prompt engineer'' which learns to query a separate neural network model for abstract reasoning, planning and decision making. Neural network distillation~\citep{schmidhuber1992} can be used to collapse this model and the prompt engineer into a single chain of thought system~\citep{schmidhuber2018one}.
Recent research demonstrates that incorporating task decomposition during LLM inference improves performance on language tasks.
~\citet{wei2022chain} showed that explicit chain-of-thought task decomposition during inference significantly enhances the capabilities of LLMs. Approaches like least-to-most prompting~\citep{zhou2022least} and ReAct~\citep{yao2022react} explicitly interleave task execution and decomposition, while ReasonFlux~\citep{yang2025reasonflux} proposed a template-based method for generating reasoning trajectories. For long-form writing, flat planning methods face challenges, as the complex hierarchical dependencies within linear context history can become unwieldy and lead to a loss of coherence.
Other works have explored hierarchical decomposition approaches. For example,~\citet{khot2023decomposed} designed a modular planner-executor system with distinct few-shot prompts that can recursively decompose tasks into smaller problems of the same form. ADaPT~\citep{prasad2023adapt} introduced on-demand recursive decomposition, yet did not address the integration of fundamentally different types of operations such as retrieval and reasoning. These existing methods primarily focused on the reasoning tasks. In this work, we propose a heterogeneous recursive framework that effectively handles long-form writing tasks with distinct operational characteristics.
Our goal-decomposition approach is also distinct from and complementary to path exploration methods like ToT~\citep{yao2023tree}, CoR~\citep{wang2025cor}. Whereas these methods explore multiple parallel reasoning paths to optimize a single step, our framework focuses on decomposing a complex primary goal into a structured hierarchy of executable sub-tasks. 

\paragraph{Agent workflow.} Agent workflow defines and control the execution logic between sub-modules in an agent system. Several frameworks have been proposed to model multi-agent workflows. 
MetaGPT~\citep{hong2023metagpt} employs a standardized operating procedure for workflow representation, simplifying agent orchestration. GPTSwarm~\citep{zhuge2024gptswarm} constructs agents using graphs.
StateFlow~\citep{wu2024stateflow} models workflows as finite state machines, where each task-solving step corresponds to a state with associated output functions, though the methodology for defining states remains unspecified. While IoA~\citep{chen2024internet}'s Internet-inspired architecture enables multi-device collaboration, it does not address the coordination of cognitive tasks. Recent work has explored search-based optimization of agent workflows~\citep{sordoni2023joint,khattab2024dspy,zhuge2024gptswarm}. For example, AFlow~\citep{zhang2024aflow} optimizes workflow represented as interconnected action nodes using Monte Carlo Tree Search (MCTS). However task specific optimized workflows remain fixed rather than dynamically adapting to different inputs. This limitation becomes particularly apparent in complex scenarios like long-form writing, where agents need to flexibly alternate between different types of operations based on dynamic context.

\section{LongReport with Human Evaluations}
\label{apsec:human}
This section investigates our framework's capabilities in extended report writing. The complete dataset, including all generated reports, is publicly available at \url{https://github.com/principia-ai/WriteHERE/blob/main/test_data/examples}.

\subsection{LongReport Dataset}
Generating long-form reports presents significant challenges to a model's writing proficiency, content organization, and overall compositional skills. Furthermore, effective evaluation in this domain necessitates detailed instructions to specify report content. The existing WildSeek dataset does not adequately meet these requirements, as its prompts are relatively concise and lack sufficient detail to describe user intent. Additionally, this work seeks to establish more fine-grained distinctions among the capability dimensions emphasized across different thematic domains of long-form reports. To address these limitations, we designed a new dataset, LongReport, comprising 12 samples. These samples were crafted based on an analysis of trends, technologies, and terminology current as of April 2025.

The LongReport dataset is designed to comprehensively assess the advanced capabilities of models in producing detailed, analytical, and well-structured long-form reports. It evaluates a model's proficiency across a spectrum of complex cognitive tasks.

The dataset is meticulously organized into three core categories:
\begin{itemize}
    \item Complex Information Retrieval: This category evaluates the model's capacity to locate, filter, and initially organize scattered, ambiguous, rapidly evolving, or highly specialized information.
    \item Analysis and Information Integration: This focuses on the model's skill in dissecting diverse information, identifying intrinsic connections, conducting comparative analyses, discovering trends, and synthesizing a holistic understanding.
    \item High-Quality In-Depth Long-Form Writing: This assesses the model's ability to construct reports with a robust structure, insightful argumentation, clear expression, and persuasive content, often tackling complex socioeconomic impacts or ethical deliberations.
\end{itemize}
Topics within each category are tiered by difficulty. They are frequently situated in scenarios reflecting the near past or contemporary landscape (e.g., conditions prevalent around early to mid-2025), demanding sophisticated interpretation of emerging signals, evolving data, and the extraction of substantive insights from potentially limited, ambiguous, or marketing-oriented sources. The detailed contents of this dataset are shown in Table~\ref{aptab:dataset_1} and Table~\ref{aptab:dataset_2}.

\begin{table}[t]
\centering
\caption{Statistics of the collected reports.}
\label{aptab:statistics}
\small
\begin{tabular}{lcccccc}
\toprule
& \multicolumn{3}{c}{\textbf{Gemini-DR}} & \multicolumn{3}{c}{\textbf{WriteHERE}} \\
\cmidrule(lr){2-4} \cmidrule(lr){5-7} 
\textbf{ID} & Word Counts & \# Sections & \# Pages & Word Counts & \# Sections & \# Pages \\
\midrule
1  & 15,528 & 6  & 37 & 16,896 & 4  & 43  \\
2  & 14,746 & 4  & 35 & 37,133 & 9  & 85  \\
3  & 12,383 & 10 & 32 & 18,385 & 7  & 41  \\
4  & 15,355 & 7  & 37 & 28,413 & 7  & 61  \\
5  & 20,790 & 8  & 46 & 17,274 & 9  & 32  \\
6  & 16,516 & 6  & 35 & 48,437 & 11 & 106 \\
7  & 18,813 & 4  & 42 & 36,868 & 9  & 81  \\
8  & 12,942 & 3  & 27 & 26,285 & 8  & 54  \\
9  & 17,001 & 5  & 46 & 23,880 & 10 & 53  \\
10 & 18,145 & 6  & 42 & 24,942 & 7  & 51  \\
11 & 19,570 & 6  & 46 & 15,790 & 6  & 31  \\
12 & 12,301 & 5  & 28 & 21,544 & 6  & 44  \\
\bottomrule
\end{tabular}
\end{table}

\subsection{Experiment Setting}
We evaluated our model against Gemini Deep-Research with 2.5 Pro, which is recognized as one of the state-of-the-art report writing models, as a strong baseline. We designed a pairwise evaluation comparing the reports generated by WriteHERE (Gemini 2.5 Pro) and Gemini Deep-Research (2.5 Pro) on the same topics from the LongReport dataset. For WriteHERE, we use the same configuration as in the experiments on WildSeek. Each prompt is attached a general suffix: \emph{Write a detailed, in-depth, and comprehensive report exceeding 10,000 words.}

According to the official documentation\footnote{https://gemini.google/overview/deep-research/?hl=en}, Gemini-Deep Research represents a specialized variant that has undergone additional training beyond the foundational 2.5 Pro model, with specific optimization for report generation tasks. The architecture may incorporate a multi-model framework; however, it fundamentally comprises a dedicated model that has been fine-tuned to enhance several critical capabilities: problem decomposition during the planning phase, sub-question dependency modeling in the search phase, and synthesis with reflection mechanisms during the writing phase. In contrast, our methodology employs Google's general-purpose foundation model, Gemini-2.5-pro-preview-05-06, and harnesses its inherent capabilities through the implementation of the WriteHERE framework.

\subsection{Human Evaluation}
For the human evaluation phase, five volunteers were recruited. These individuals were neither students nor members of our laboratory to ensure an external perspective. To ensure a thorough understanding of the report content, all annotators possessed at least a Bachelor's degree, a necessary qualification due to the technical nature of the evaluation task. Annotators were instructed to evaluate the generated reports based on detailed guidelines, which are provided below. In the informed consent form, we clarify to participants that while the results of the research study may be presented at scientific or professional meetings or published in scientific journals, their identity will remain confidential and will not be disclosed at any point. Compensation for the annotators was determined based on the complexity of the evaluation tasks and the expertise required. 

\paragraph{Evaluation criteria.} For the evaluation criteria, we followed the original 4 dimensions proposed by~\citet{jiang2024into} as used in WildSeek. For long report evaluation, we added one additional dimension—Clarity, Cohesion, and Language—to assess organization and language use. We showed evaluators the original prompt along with two reports generated from the same prompt, asking them to assign rubric scores from 1-5 to each dimension for each report, and then select which report they overall preferred. Furthermore, we instructed the evaluators to minimize the impact of formatting elements on their assessment and concentrate on the actual content of the reports. The details are shown in Section~\ref{apsubsec:criteria}. 

\paragraph{Evaluation setup.} Each evaluator assessed all 12 pairs of reports, resulting in 5 reference scores per report. The data presented to evaluators was randomly shuffled in order, and the arrangement within each group was also randomly shuffled to eliminate order bias. File names are anonymous. Evaluators were unaware of the source of the reports in the dataset, knowing only that they were AI-generated, and had no knowledge of how many different AI models were involved.

\begin{table}[t]
\centering
\caption{Human evaluation results. "Overall" denotes the overall vote counts in the pairwise comparison.}
\label{aptab:results}
\small
\begin{tabular}{lllcccccc}
\toprule
\multicolumn{9}{c}{\textit{\textbf{Category One: Complex Information Retrieval}}} \\
\midrule
\textbf{Level} & \textbf{ID} & \textbf{Method} & Relevance & Breadth & Depth & Novelty & Clarity & Overall \\
\midrule
\multirow{4}{*}{1} & \multirow{2}{*}{1} & Gemini-DR & 4.8 & 4.6 & 3.8 & 4.2 & 4.8 & 5 \\
                   &                    & WriteHERE   & 4.6 & 4.4 & 3.4 & 3.6 & 4.0 & 0 \\
\cmidrule(lr){2-9}
                   & \multirow{2}{*}{2} & Gemini-DR & 4.8 & 4.2 & 4.0 & 3.6 & 4.4 & 3 \\
                   &                    & WriteHERE   & 5.0 & 4.4 & 4.2 & 3.8 & 3.8 & 2 \\
\midrule
\multirow{4}{*}{2} & \multirow{2}{*}{3} & Gemini-DR & 4.0 & 4.4 & 4.2 & 3.8 & 4.2 & 2 \\
                   &                    & WriteHERE   & 4.6 & 4.4 & 4.4 & 4.2 & 4.4 & 3 \\
\cmidrule(lr){2-9}
                   & \multirow{2}{*}{4} & Gemini-DR & 4.6 & 3.8 & 3.6 & 3.8 & 3.8 & 2 \\
                   &                    & WriteHERE   & 4.2 & 4.4 & 4.0 & 3.6 & 4.2 & 3 \\
\midrule
\multicolumn{9}{c}{\textit{\textbf{Category Two: Analysis and Information Integration}}} \\
\midrule
\multirow{4}{*}{1} & \multirow{2}{*}{5} & Gemini-DR & 4.8 & 4.0 & 3.6 & 3.8 & 4.0 & 2 \\
                   &                    & WriteHERE   & 4.8 & 4.6 & 4.2 & 4.0 & 4.6 & 3 \\
\cmidrule(lr){2-9}
                   & \multirow{2}{*}{6} & Gemini-DR & 4.6 & 4.2 & 3.4 & 3.4 & 4.4 & 3 \\
                   &                    & WriteHERE   & 4.6 & 4.6 & 4.2 & 3.8 & 4.6 & 2 \\
\midrule
\multirow{4}{*}{2} & \multirow{2}{*}{7} & Gemini-DR & 4.0 & 4.0 & 3.6 & 3.2 & 4.2 & 1 \\
                   &                    & WriteHERE   & 5.0 & 4.6 & 4.6 & 4.0 & 4.4 & 4 \\
\cmidrule(lr){2-9}
                   & \multirow{2}{*}{8} & Gemini-DR & 4.4 & 4.0 & 3.8 & 3.6 & 4.4 & 2 \\
                   &                    & WriteHERE   & 4.0 & 4.2 & 3.6 & 4.0 & 4.4 & 3 \\
\midrule
\multicolumn{9}{c}{\textit{\textbf{Category Three: High-Quality In-Depth Long-Form Writing}}} \\
\midrule
\multirow{4}{*}{1} & \multirow{2}{*}{9} & Gemini-DR & 4.4 & 4.0 & 4.2 & 3.8 & 3.8 & 2 \\
                   &                    & WriteHERE   & 4.4 & 4.6 & 4.0 & 4.4 & 4.2 & 3 \\
\cmidrule(lr){2-9}
                   & \multirow{2}{*}{10} & Gemini-DR & 4.0 & 4.2 & 3.8 & 3.8 & 4.6 & 3 \\
                   &                     & WriteHERE   & 4.0 & 4.4 & 3.6 & 4.0 & 3.8 & 2 \\
\midrule
\multirow{4}{*}{2} & \multirow{2}{*}{11} & Gemini-DR & 4.8 & 4.6 & 4.4 & 4.0 & 4.2 & 3 \\
                   &                     & WriteHERE   & 4.6 & 4.6 & 3.8 & 3.8 & 4.6 & 2 \\
\cmidrule(lr){2-9}
                   & \multirow{2}{*}{12} & Gemini-DR & 4.4 & 4.6 & 3.6 & 4.0 & 4.4 & 2 \\
                   &                     & WriteHERE   & 4.4 & 4.4 & 4.2 & 4.4 & 4.2 & 3 \\
\midrule 
\multirow{2}{*}{\textbf{Overall}} &       & \textbf{Gemini-DR} & 4.5 & 4.2 & 3.8 & 3.8 & 4.3 & 5 \\ 
                                 &       & \textbf{WriteHERE}   & 4.5 & 4.5 & 4.0 & 4.0 & 4.3 & 7 \\ 
\bottomrule
\end{tabular}
\end{table}

\paragraph{Data preprocessing.} We implemented several preprocessing procedures to ensure that articles generated by both methods were as similar as possible in format, thereby minimizing potential bias from formatting differences in content evaluation. First, we removed all article titles, as our method did not explicitly instruct the model to generate titles, and generating appropriate titles for given reports is a relatively secondary and straightforward task. Second, we removed appendices while retaining only citation markers, due to the difficulty of standardizing formats and our current focus not including the evaluation of citation source fidelity, which represents a dimension relatively independent of long-form writing. We used Microsoft Word to maintain consistency in citation markers and thematic style throughout the articles to eliminate stylistic influences. However, it should be noted that some stylistic factors proved difficult to eliminate: for instance, our observations indicate that Gemini-Deep Research generated articles typically contain extensive tables and, in most cases, include an Executive Summary at the beginning of the report, whereas our method did not specify the prior generation of an Executive Summary. Furthermore, given the potential for outline adjustments during the writing process, generating a summary at the beginning would be inappropriate. The detailed statistics of the reports are shown in Table~\ref{aptab:statistics}.

\subsection{Evaluation Results}
The full results is shown in Table~\ref{aptab:results}. According to the results, our method in general generates articles comparable to those produced by Gemini Deep Research. In average scores, WriteHERE demonstrates a slight advantage in breadth, depth, and novelty, achieving a 7:5 overall vote. 
A detailed review of the evaluation results for each sample highlighted a generally balanced performance between the two systems. Most overall votes clustered between 2 and 3, underscoring the difficulty in definitively distinguishing a superior method in many instances. However, Samples 1 and 7 presented notable exceptions, where the performance disparity was more pronounced.

Interestingly, the study found that article length did not significantly sway annotators' scores. For example, in Sample 6, WriteHERE generated a substantial 48,000-word article yet received a lower vote score than Gemini. This suggests that reviewers diligently adhered to the specific evaluation criteria, rather than being influenced by output volume.

A closer analysis of Sample 7 indicates that Gemini Deep Research did not complete its planned content, which likely contributed to its lower scores in breadth and depth. In samples where Gemini outperformed, it typically scored higher in clarity, possibly due to superior content organization. Conversely, where our method prevailed, it generally excelled in breadth, depth, and novelty.

According to its official documentation, Gemini critically evaluates information, identifies key themes and inconsistencies, and structures reports logically and informatively, incorporating multiple self-critique rounds to enhance clarity and detail. The Deep Research feature specifically employs iterative self-reflection mechanisms to optimize article clarity, with training designed to strengthen these capabilities. These documented characteristics of Gemini align with the evaluation results observed in our study.

\subsection{Evaluation Criteria}
\label{apsubsec:criteria}
\subsubsection*{Thoroughness of Coverage (Breadth)}
\textit{How completely does the report cover all important aspects of the topic?}

\begin{tabular}{cL{14cm}}
\toprule
\textbf{Score} & \textbf{What to Look For} \\
\midrule
1 & \textbf{Minimal Coverage}: The report barely touches on the topic, missing most key elements. You'll notice major gaps that prevent basic understanding. \\
2 & \textbf{Limited Coverage}: The report includes some important aspects but leaves out several critical elements. The picture feels incomplete. \\
3 & \textbf{Adequate Coverage}: Most main aspects are covered, though you might find some relevant points missing or notice unnecessary details included. \\
4 & \textbf{Comprehensive Coverage}: All major points are addressed with appropriate detail and minimal irrelevant information. The coverage feels well-balanced. \\
5 & \textbf{Exceptional Coverage}: The report thoroughly examines all important aspects with ideal depth, excluding anything irrelevant. Nothing significant is missing. \\
\bottomrule
\end{tabular}

\subsubsection*{Innovative Content (Novelty)}
\textit{Does the report go beyond the obvious to include valuable related information?}

\begin{tabular}{cL{14cm}}
\toprule
\textbf{Score} & \textbf{What to Look For} \\
\midrule
1 & \textbf{No Innovation}: The report strictly follows predictable content with nothing added beyond what was directly requested. \\
2 & \textbf{Minimal Innovation}: Contains a few new angles or insights, but they add little value to the overall understanding. \\
3 & \textbf{Moderate Innovation}: Introduces some fresh perspectives or related information that somewhat enhances the report. \\
4 & \textbf{Good Innovation}: Includes several valuable new aspects that meaningfully expand on the requested information. \\
5 & \textbf{Outstanding Innovation}: Presents numerous highly relevant additional insights that significantly enrich understanding while remaining connected to the core topic. \\
\bottomrule
\end{tabular}

\subsubsection*{Focus and Relevance (Relevance)}
\textit{Does the report stay on topic and deliver what was requested?}

\begin{tabular}{cL{14cm}}
\toprule
\textbf{Score} & \textbf{What to Look For} \\
\midrule
1 & \textbf{Unfocused}: The report wanders significantly off-topic, containing much irrelevant material that doesn't serve the purpose. \\
2 & \textbf{Poorly Focused}: Contains some relevant content, but frequently drifts into tangential or unrelated areas. \\
3 & \textbf{Moderately Focused}: Mostly stays on topic with occasional diversions that still provide some useful information. \\
4 & \textbf{Well-Focused}: Maintains clear relevance throughout with only minor deviations that add value to the core topic. \\
5 & \textbf{Laser-Focused}: Perfectly addresses the request with every element clearly contributing to the purpose, even when exploring related aspects. \\
\bottomrule
\end{tabular}

\subsubsection*{Depth of Analysis (Depth)}
\textit{How thoroughly does the report explore the topic beneath the surface?}

\begin{tabular}{cL{14cm}}
\toprule
\textbf{Score} & \textbf{What to Look For} \\
\midrule
1 & \textbf{Very Shallow}: Offers only basic facts or observations without meaningful explanation or analysis. \\
2 & \textbf{Shallow}: Provides some explanation but fails to explore important complexities or implications. \\
3 & \textbf{Moderate Depth}: Examines key aspects with some analysis, though certain important areas lack detailed exploration. \\
4 & \textbf{Substantial Depth}: Explores most aspects thoroughly with good analysis of complexities and interconnections. \\
5 & \textbf{Exceptional Depth}: Provides comprehensive analysis of all relevant aspects, revealing nuances, underlying factors, and broader significance. \\
\bottomrule
\end{tabular}

\subsubsection*{Clarity, Cohesion, and Language (Clarity)}
\textit{How clear, well-organized, and grammatically sound is the report's language and structure?}

\begin{tabular}{cL{14cm}}
\toprule
\textbf{Score} & \textbf{What to Look For} \\
\midrule
1 & \textbf{Poor}: The report is very difficult to understand due to pervasive errors in grammar, spelling, or punctuation. Language is frequently ambiguous or incorrect. Structure is chaotic, lacking logical flow between sentences and paragraphs, severely hindering comprehension. \\
2 & \textbf{Problematic}: Significant issues with clarity, cohesion, or language make the report challenging to read. Frequent errors, awkward phrasing, or inconsistent terminology are common. The structure may be weak, with poor transitions and organization that obscure the main points. \\
3 & \textbf{Acceptable}: The report is generally understandable, but contains noticeable errors in grammar, spelling, or word choice. Clarity or cohesion may falter in places, with some awkward sentences or less-than-smooth transitions. The overall structure is present but could be significantly improved. \\
4 & \textbf{Good}: The report is clearly written and well-organized. Language is precise and appropriate, with minimal errors in grammar, spelling, or punctuation that do not impede understanding. Sentences and paragraphs flow logically with effective transitions. \\
5 & \textbf{Excellent}: The report demonstrates exceptional clarity, precision, and fluency. Language is sophisticated, engaging, and virtually error-free. The structure is highly effective, with seamless cohesion and logical progression of ideas that enhance readability and impact. \\
\bottomrule
\end{tabular}

\begin{table}[t]
\centering
\caption{LongReport Dataset: Category 1 and 2.}
\label{aptab:dataset_1}
\small
\begin{tabular}{|c|c|p{1.3cm}|p{8.8cm}|p{2.5cm}|}
\hline
\multicolumn{5}{|c|}{\textit{\textbf{Category One: Complex Information Retrieval}}} \\
\hline
\textbf{Level} & \textbf{ID} & \textbf{Topic} & \textbf{Prompt} & \textbf{Remark} \\
\hline

\multirow{2}{*}{1} & 1 & Finance & 
Evaluate the preliminary commercial viability and market adoption of decentralized finance (DeFi) protocols specifically designed for the tokenization of real-world assets (RWA) as of Q2 2025. Focus on analyzing publicly available on-chain data, early user feedback, project whitepapers, and the sustainability of their economic models beyond the experimental phase. &
Information has relatively clear tracking channels such as project announcements, on-chain explorers, community forums, etc. \\
\cline{2-5}

& 2 & Commerce & 
Investigate and analyze potential structural adjustments or "invisible" restructuring of critical strategic minerals (such as rare earths, lithium, cobalt, etc.) supply chains that may be occurring but not widely reported as of April 2025, against the backdrop of evolving global geopolitical dynamics. Focus on identifying and interpreting these early signals based on international trade data, policy trends in major countries, corporate investment announcements, logistics hub changes, and industry expert interviews (assuming second-hand summaries are available). &
Information sources are diverse, requiring careful screening and correlation analysis. \\
\hline

\multirow{2}{*}{2} & 3 & Technology & 
Conduct in-depth research and critically evaluate the autonomous decision-making capabilities demonstrated by enterprise-level Agentic AI systems in public demonstrations or early pilots as of Q1 2025. Focus on discerning whether they have truly moved toward autonomous operation or remain highly complex automation of preset processes, and attempt to identify key "human intervention" nodes or implicit dependencies based on publicly disclosed technical information and expert interpretations. &
Requires extremely strong discernment ability and understanding of underlying technical logic. \\
\cline{2-5}

& 4 & Technology & 
Investigate and analyze whether, beyond widely known major announcements, there are more subtle or non-explicitly claimed signals of quantum advantage/supremacy for specific narrow problems from research institutions or private enterprises between late 2024 and early 2025. Focus on interpreting technical preprints, small-scale discussions among domain experts, and preliminary reports from professional conferences that might suggest such breakthroughs. &
Information acquisition is extremely difficult, requiring deep access to specific academic circles or unconventional information sources. \\
\hline

\multicolumn{5}{|c|}{\textit{\textbf{Category Two: Analysis and Information Integration}}} \\
\hline
\textbf{Level} & \textbf{ID} & \textbf{Topic} & \textbf{Prompt} & \textbf{Remark} \\
\hline

\multirow{2}{*}{1} & 5 & Business & 
Analyze the specific impacts of integrated generative AI on corporate ESG (Environmental, Social, Governance) reporting practices as of mid-2025. Focus on examining measurable progress in data coordination efficiency improvement, generation of real-time sustainability insights, and enhancement of corporate information disclosure transparency, integrating public case studies, industry research data, and reports published by companies themselves. &
Information sources are relatively easy to obtain, with emphasis on summarization and synthesis. \\
\cline{2-5}

& 6 & Policy & 
Conduct a comparative analysis of core laws, regulations, regulatory frameworks, and enforcement practices regarding cross-border data flows in major global economies (such as the US, EU, China, India, etc.) as of Q1 2025. Focus on systematically analyzing similarities and differences in data localization requirements, personal information protection standards, data security review mechanisms, and international data transfer protocols, and assess the specific impacts of these differences on multinational corporations' global operation strategies, compliance costs, and innovation activities. &
Information mainly comes from official documents and professional interpretations, but requires structured integration and in-depth analysis. \\
\hline

\multirow{2}{*}{2} & 7 & Finance & 
Comprehensively analyze investment trends and key innovation directions in the climate technology field as of mid-2025. Integrate public financing data, patent applications, scientific papers, and technology progress reports for AI-optimized renewable energy, carbon capture and storage technologies, sustainable agriculture, and emerging clean technology business models, and assess their collective progress and bottlenecks in driving global industrial decarbonization goals. &
Requires processing and correlating large amounts of different types of data and distilling trends and bottlenecks from them. \\
\cline{2-5}

& 8 & Cyber-security & 
Review the evolution and deployment status of digital trust architectures and related frameworks (such as content provenance standards, identity verification technologies, media authenticity detection tools) in response to increasingly complex synthetic media challenges (such as deepfakes, false information dissemination) as of mid-2025. Focus on analyzing technological advances, industry standard-setting, policy and regulatory responses, and actual application effects and limitations in data provenance, multimodal identity verification, and transaction verification resilience. &
Information is scattered across different resources, requiring high-level integration and forward-looking analysis. \\
\hline
\end{tabular}
\end{table}

\clearpage

\begin{table}[h]
\centering
\caption{LongReport Dataset: Category 3.}
\label{aptab:dataset_2}
\small
\begin{tabular}{|c|c|p{1.3cm}|p{8.8cm}|p{2.3cm}|}
\hline
\multicolumn{5}{|c|}{\textit{\textbf{Category Three: High-Quality In-Depth Long-Form Writing}}} \\
\hline
\textbf{Level} & \textbf{ID} & \textbf{Topic} & \textbf{Prompt} & \textbf{Remark} \\
\hline

\multirow{2}{*}{1} & 9 & Policy & 
Based on the understanding level of 2025, write a comprehensive report outlining strategic pathways for significantly expanding the scale of regenerative agriculture practices and effectively promoting livestock methane emission reduction technologies by 2030. The report should deeply explore multiple dimensions including economic feasibility, policy incentive mechanisms, technological maturity and promotion, challenges in farmer adoption willingness, and consumer awareness enhancement, and provide actionable recommendations. &
Although complex, each sub-field already has a certain foundation of research and discussion. \\
\cline{2-5}

& 10 & Technology & 
Present an engaging narrative report showcasing how various technologies (such as AI, IoT, biotechnology, clean energy technology, etc.) are being innovatively applied by startups, research institutions, and social enterprises to advance specific United Nations Sustainable Development Goals (SDGs) (e.g., clean water and sanitation, affordable and clean energy, responsible consumption and production) as of mid-2025. The report should include influential case analyses, analyzing success factors, challenges faced, and scalability, and explore the broader ecosystem supporting "Tech for Good" (policies, investments, collaborations). &
Emphasis on narrative ability and depth of case analysis. \\
\hline

\multirow{2}{*}{2} & 11 & Technology & 
Write a thoughtful exploratory report examining how Ambient Intelligence (AmI) systems (such as smart homes, smart city infrastructure, personalized medical monitoring) are redefining human living and working spaces as of 2025. Critically examine the balance between convenience and efficiency brought by passive sensing and predictive automation, and deeper ethical considerations such as privacy invasion, data misuse, algorithmic bias, digital divide, weakening of human autonomy, and social control, and propose guiding principles and governance frameworks for building "human-centered" ethical intelligent spaces. &
Requiring detailed and balanced argumentation and constructive governance thinking. \\
\cline{2-5}

& 12 & Futures Research & 
Write a forward-looking report exploring the evolutionary pattern of Hybrid Autonomous Systems as of April 2025, particularly the integration of human supervision and AI decision-making in key areas such as infrastructure management (e.g., smart grids, autonomous driving traffic networks), healthcare (e.g., AI-assisted diagnosis and surgery), and scientific discovery. Conduct in-depth analysis of best practice models for human-machine collaboration, new challenges in building trust, defining responsibilities, and skill gaps, as well as the profound impact and necessary adjustments this collaborative model will have on future skill requirements, education systems, labor market structures, and even social equity. &
Requires high insight, foresight, and comprehensive grasp of complex system effects. \\
\hline
\end{tabular}
\end{table}

\section{Experiments Details}
\label{apsec:details}
In this section we introduce the implementation details of our experiments. We also provide additional experiments for the evaluation results. Scores on WildSeek in this section is produced with the open-sourced evaluator LLM Prometheus 2~\citep{kim2024prometheus} which is shown to have high agreement with the proprietary LM judges.

\paragraph{General configuration.} For the base LLMs in all the experiments, we employed GPT-4o-20240806, Claude-3.5-Sonnet-20241022, DeepSeek-R1 with their default parameters.

\subsection{Topic template for WildSeek}
In the WildSeek dataset, each sample contains two key fields: Topic and Intent. The Co-STORM paper~\citep{jiang2024into} implements different experimental approaches for these fields. Specifically, for Co-STORM, both the Topic and Intent fields are combined and provided to a LLM that simulates user behavior. In contrast, when using STORM, only the Topic field is supplied to the model.
In our implementation we combine the Topic and Intent fields into a refined topic before feeding them to the agents. We remove the trailing period or question mark from the Topic field and make the first letter of the Intent field lowercase. The final refined topic is created with the following template: \texttt{f"\{topic\}, \{intent\}"}.

\subsection{STORM}
\label{sec:storm_variants}

\begin{table*}[htb]
\centering
\begin{tabular}{lllc cccc}
\toprule
\multirow{2}{*}{\textbf{Backbone}} & \multirow{2}{*}{\textbf{Method}} & \multirow{2}{*}{\textbf{Tag or branch}} & \multicolumn{4}{c}{\textbf{Report Quality}} \\
\cmidrule(lr){4-7}
& & & \textbf{Relevance} & \textbf{Breadth} & \textbf{Depth} & \textbf{Novelty} \\
\midrule
\multirow{2}{*}{GPT-4o} & \multirow{2}{*}{STORM}
& \small{\texttt{v1.1.0}} & 4.500  & 4.530 & 4.693 & 4.214 \\
& & \small{\texttt{NAACL-2024-code-backup}}
& 4.580\textsuperscript{\textcolor{green}{+0.080}}  
& 4.320\textsuperscript{\textcolor{red}{-0.210}}  
& 4.617\textsuperscript{\textcolor{red}{-0.076}}  
& 3.913\textsuperscript{\textcolor{red}{-0.301}} \\
\bottomrule
\end{tabular}
\caption{Reproduction experiments for STORM with GPT-4o backbone and Serp/Bing retriever.}
\label{tab:storm_naacl_vs_v1p1}
\end{table*}

Running STORM baseline from the official release branch \texttt{NAACL-2024-code-backup} would involve quering multiple outdated LLMs, specifically, \texttt{gpt-3.5-turbo}, \texttt{gpt-4}, and \texttt{gpt-4-32k}. We use \texttt{gpt-4o} for the fair comparison of the orchestration-level algorithms. We compare the \texttt{NAACL-2024-code-backup} branch to \texttt{v1.1.0} tag of the official STORM repository in Table \ref{tab:storm_naacl_vs_v1p1}. We observe that the most recent code \texttt{v1.1.0} is slightly stronger than the official branch \texttt{NAACL-2024-code-backup} on average across rubrics. We choose tag \texttt{v1.1.0} as a stronger baseline. We follow the default hyperparameter setting in the official implementation.

\subsection{Co-STORM}
\label{apsec:costorm_variants}

\begin{table*}[htb]
\centering
\begin{tabular}{lllc cccc}
\toprule
\multirow{2}{*}{\textbf{Backbone}} & \multirow{2}{*}{\textbf{Method}} & \multirow{2}{*}{\textbf{Input}} & \multirow{2}{*}{\textbf{Variant}} & \multicolumn{4}{c}{\textbf{Report Quality}} \\
\cmidrule(lr){5-8}
& & & & \textbf{Relevance} & \textbf{Breadth} & \textbf{Depth} & \textbf{Novelty} \\
\midrule
\multirow{3}{*}{GPT-4o} & \multirow{3}{*}{Co-STORM}
& T & \texttt{2+I+4-turn}
& 4.429 & 4.469 & 4.531 & 4.255 \\
& & T+I & \texttt{3-turn}
& 4.263\textsuperscript{\textcolor{red}{-0.166}}
& 4.384\textsuperscript{\textcolor{red}{-0.085}}
& 4.535\textsuperscript{\textcolor{green}{+0.004}}
& 3.869\textsuperscript{\textcolor{red}{-0.386}} \\
& & T+I & \texttt{1-turn}
& 4.310\textsuperscript{\textcolor{red}{-0.119}}
& 4.440\textsuperscript{\textcolor{red}{-0.029}}
& 4.380\textsuperscript{\textcolor{red}{-0.151}}
& 4.000\textsuperscript{\textcolor{red}{-0.255}} \\
\bottomrule
\end{tabular}
\caption{Performance of Co-STORM variants in terms of report quality. The input format `T' refers to that we only include Topic as the original input. `T+I' denotes we combine Topic and Intent as in STORM.}
\label{tab:costorm_varieties}
\end{table*}

The user collaborative part of Co-STORM is simulated by a LLM. We follow the example implementation provided in the official repository\footnote{https://github.com/stanford-oval/storm/blob/main/examples/costorm\_examples/run\_costorm\_gpt.py}. The user utterance simulation is executed by configuring \texttt{costorm\_runner.step(simulated\_user=True, simulate\_user\_intent=intent)}. To align with the default setting of STORM, we set \texttt{max\_search\_queries} to 3. We set the number of turns after the warm-up phase but before the simulated user utterance to 2, and the number of turns following to 4, thereby simulating 1 and 2 rounds of round-table discussions, respectively.

We conducted a comparative analysis of different Co-STORM variants, as presented in Table~\ref{tab:costorm_varieties}. The implementation described previously is denoted as $\texttt{2+I+4-turn}$. We implemented two variants that use combined Topic and Intent as input, consistent with the approach in both STORM and our method. The two variants differ in the number of turns following the warm-up phase. The first variant employs 3 turns after the warm-up phase without simulated user utterance (designated as "3-turn" in the table). This configuration adheres to the default settings specified in the official example. The other uses just 1 turn (labeled as "1-turn" in the table).
The results show that they are relatively worse than the $\texttt{2+I+4-turn}$ variant, especially in the novelty dimension. We thus present the results of the $\texttt{2+I+4-turn}$ variant in the main paper.

\subsection{Search Agent}
\label{apsec:s_agent}
The search agent implementation in STORM and Co-STORM follows a retrieval-augmented generation approach but differs in their information seeking strategies. In STORM, the search agent converts questions into multiple search queries using an LLM, retrieves results through search APIs, and applies rule-based filtering following Wikipedia's reliable sources guidelines to exclude unreliable sources like social media posts and personal blogs. Co-STORM extends this with a multi-perspective search strategy where agents with different expertise generate questions based on their specialized viewpoints. It also implements a dynamic reranking mechanism that scores retrieved information using a formula which prioritizes information that is relevant to the topic but not directly answering the original question. In our implementation, we employ a multi-agent system as the search agent, consisting of a ReAct-style retrieval agent, a result ranking agent, and a content summarization agent. The retrieval agent issues up to 4 queries and retrieves a maximum of 20 results. These results are then passed to the ranking agent, which scores them and selects the top four based on relevance. The content summarization agent then extracts information from these top-ranked results that is most relevant to the query and search intent, before returning them to the upper-level search execution process. The cost-efficient model \texttt{gpt-4o-mini} is used for the ranking and summarization stages.

\section{A Detailed Walkthrough of the Proposed Framework}
\label{apsec:walk}
This appendix provides a concrete example of how the proposed framework dynamically plans and executes a complex, long-form writing task. We trace the evolution of the task graph through several key ``snapshots'' to illustrate our framework working process.

\subsection{The Initial Task}
The process begins with a single, high-level user goal, which becomes the \textsc{Active} root node of our task graph.

\begin{tcolorbox}[colback=black!5!white,colframe=black!75!black,title=Root Task (ID: 0)]
\textbf{Type:} \texttt{write} \\
\textbf{Goal:} Comprehensively analyze investment trends and key innovation directions in the climate technology field as of mid-2025. Integrate public financing data, patent applications, scientific papers, and technology progress reports for AI-optimized renewable energy, carbon capture and storage technologies, sustainable agriculture, and emerging clean technology business models, and assess their collective progress and bottlenecks in driving global industrial decarbonization goals.\\
\textbf{Dependencies:} None \\
\textbf{Result:} None
\end{tcolorbox}

\subsection{Execution Snapshots}
\subsubsection{Snapshot 1: Initial Planning for the Root Task}
The process begins with the scheduler selecting the \textsc{Active} root node. The ``Atomicity Determination'' module judges the task as complex and non-primitive, thus invoking the \texttt{TypedPlan} function for decomposition.

\begin{itemize}[leftmargin=*]
    \item \textbf{Scheduler Action:} Selects \texttt{Root (ID: 0)}.
    \item \textbf{Agent Action:} Decomposes the root goal into a sequence of high-level steps: an initial search, a thinking/outlining phase, and three major writing phases.
    \item \textbf{Graph Update:}
        \begin{itemize}
            \item The \texttt{Root} node's state is updated to \textsc{Suspend}.
            \item Five child nodes are added to the graph:
                \begin{itemize}
                    \item \texttt{1 [search]}: Collect overview data on global climate tech investment trends, key policy drivers, and identify major sources/databases for public financing, patent applications, scientific papers, and technology progress reports relevant up to mid-2025 to inform the overall report structure and initial analysis.
                    \item \texttt{2 [think]}: Based on the initial search (Task 1) and the overall report goal, design a detailed report outline. This includes defining the analytical framework for assessing investment trends, innovation directions, progress, and bottlenecks for each specified technology area (AI-optimized renewable energy, CCS, sustainable agriculture, emerging clean tech business models) and for their collective impact. Specify key metrics and data points to be extracted from public financing data, patent applications, scientific papers, and technology progress reports for the mid-2025 timeframe
                    \item \texttt{3 [write]}: Write the introductory sections of the report, including the main Introduction and the chapter on the Overall Climate Technology Landscape as of Mid-2025. These sections will set the context, define scope and methodology, and provide a high-level overview of the climate tech sector.
                    \item \texttt{4 [write]}: Write the detailed analysis chapters for the four key climate technology areas: AI-Optimized Renewable Energy, Carbon Capture and Storage (CCS) Technologies, Sustainable Agriculture, and Emerging Clean Technology Business Models, as of mid-2025. Each analysis will comprehensively cover investment trends (integrating public financing data), key innovation directions (from patent applications, scientific papers, tech progress reports), progress towards decarbonization, and identified bottlenecks for each area.
                    \item \texttt{5 [write]}: Write the concluding sections of the report, including the Cross-Cutting Analysis and Assessment, and the overall Conclusion and Outlook, as of mid-2025. These sections will synthesize the findings from the technology deep dives and provide a holistic view of the climate tech field's contribution to industrial decarbonization.
                \end{itemize}
            \item \texttt{Task 1} has no dependencies and is set to \textsc{Active}. Other tasks are inactive as they depend on the completion of earlier tasks.
        \end{itemize}
\end{itemize}

\subsubsection{Snapshot 2: Recursive Planning \& Dynamic Goal Refinement for \texttt{Task 3}}
After \texttt{Task 1 (search)} and \texttt{Task 2 (think)} are executed, their outputs are used to inform the next planning step, showcasing the framework's adaptive nature.

\begin{itemize}[leftmargin=*]
    \item \textbf{State of the Graph:} \texttt{Task 1} and \texttt{Task 2} are \textsc{Silent}. Their outputs are stored in memory and serve as context for subsequent planning. For clarity and brevity, the extensive outputs of these tasks are presented below in a summarized format, not as their complete machine-readable versions.
    \item \textbf{Scheduler Action:} Selects \texttt{Task 3}.
    \item \textbf{Agent Action (Atomicity Determination \& Goal Refinement):} The \texttt{IsAtomic} and \texttt{Update} module are applied, which leveraged the outputs of the completed dependencies to refine the task's objective and determined the task is non-primitive. The original high-level goal, ``Write the introductory sections of the report, including the main Introduction and the chapter on the Overall Climate Technology Landscape as of Mid-2025...'', is expanded into a highly detailed directive.
\end{itemize}

\begin{tcolorbox}[colback=green!5!white,colframe=green!50!black,breakable,title=Summarization of the output of Task 1 (Search Summary),fonttitle=\small]
\begin{itemize}[leftmargin=*,topsep=0pt,partopsep=0pt]
    \item \textbf{Overall Investment Trends:} Global energy transition investment reached a record \$2.1 trillion in 2024 (BNEF) (webpage[1]), an 11\% increase...
    \item \textit{... (other search information)}
\end{itemize}
\end{tcolorbox}

\begin{tcolorbox}[colback=orange!5!white,colframe=orange!60!black,breakable,title=Summarization of the output of Task 2 (Report Outline),fonttitle=\small]
\begin{itemize}[leftmargin=*,topsep=0pt,partopsep=0pt]
    \item \textbf{Part I: Global Context (Chapters 1-2):}
        \begin{itemize}[leftmargin=*,topsep=0pt,partopsep=0pt]
            \item \textbf{Chapter 1 (Introduction):} Establishes the report's rationale, scope, objectives, and a detailed methodology covering integrated data sources...
            \item \textbf{Chapter 2 (Global Landscape):} Provides a comprehensive overview of the mid-2025 investment and policy context, including aggregate investment flows, key policy drivers (IRA, Green Deal...), and cross-cutting innovation enablers like AI...
        \end{itemize}
    \item \textbf{Part II: Deep Dive Technology Analyses (Chapters 3-6):}
        \begin{itemize}[leftmargin=*,topsep=0pt,partopsep=0pt]
            \item A recurring analytical structure will be applied to four key areas: AI-Optimized Renewable Energy..., Carbon Capture, Utilization, and Storage (CCUS)..., Sustainable Agriculture Technologies..., and Emerging Clean Tech Business Models...
            \item Each chapter will cover investment trends, innovation directions (patents, papers), technology progress, deployment levels, and identified bottlenecks...
        \end{itemize}
    \item \textbf{Part III: Synthesis and Outlook (Chapters 7-9):}
        \begin{itemize}[leftmargin=*,topsep=0pt,partopsep=0pt]
            \item \textbf{Chapter 7 (Collective Impact):} Assesses collective progress against global goals, analyzing synergies, trade-offs, and systemic challenges like infrastructure deficits and supply chain resilience...
            \item \textbf{Chapter 8 (Future Outlook):} Presents projected trends and provides strategic recommendations for policymakers and investors...
            \item \textbf{Chapter 9 (Conclusion):} Summarizes the state of climate technology in mid-2025 and issues a call to action...
        \end{itemize}
\end{itemize}
\end{tcolorbox}

The refined goal for Task 3, shown below in a condensed format for clarity, becomes:
\begin{tcolorbox}[colback=black!5!white,colframe=black!75!black,title=Refined Goal (Task ID: 3),fonttitle=\small]
\textbf{Type:} \texttt{write} \\
\textbf{Goal:} Write the 2000-word introductory part of the report, comprising Chapter 1 (Introduction) and Chapter 2 (Global Climate Technology Landscape). Chapter 1 must cover the Report Rationale, Scope \& Objectives, and a detailed Methodology (including data sources and analytical framework). Chapter 2 must analyze the Global Investment Overview, Key Policy Drivers (e.g., IRA, Green Deal), and Cross-Cutting Innovation Enablers as of mid-2025.
The entire task must adhere to the detailed outline from Task 2 and integrate specific data points (e.g., investment figures, policy names) from Task 1.\\
\textbf{Dependencies:} 1,2 \\
\textbf{Result:} None

\vspace{1ex} 
\small\textit{(Note: This is a summary of the full, highly-detailed goal generated, which specifies every sub-section and data point. It is condensed here for illustrative purposes.)}
\end{tcolorbox}

\begin{itemize}[leftmargin=*]
    \item \textbf{Agent Action (Conditional Decomposition):} Invoke the \texttt{TypedPlan} module, which decomposes this newly refined, complex goal into two more manageable sub-tasks:
        \begin{itemize}
            \item \texttt{3.1 [write]}: Write Chapter 1 (Introduction) of the report, covering 1.1 (Report Rationale), 1.2 (Scope and Objectives), and 1.3 (Methodology), adhering to the detailed structure and content points outlined in Task 2 (Report Outline).
            \item \texttt{3.2 [write]}: Write Chapter 2 (Global Climate Technology Landscape: Investment and Policy Context (mid-2025)), covering 2.1 (Global Climate Technology Investment Overview), 2.2 (Key Policy Drivers and Regulatory Environment), and 2.3 (Cross-Cutting Innovation Enablers (Brief Overview)), adhering to the detailed structure and content points outlined in Task 2 (Report Outline) and drawing extensively upon the search results and analysis from Task 1.
        \end{itemize}
    \item \textbf{Graph Update:}
        \begin{itemize}
            \item \texttt{Task 3}'s state is updated to \textsc{Suspend}.
            \item Nodes \texttt{3.1} and \texttt{3.2} are added as children. Both are set to \textsc{Active}.
        \end{itemize}
\end{itemize}

\subsubsection{Snapshot 3: Atomic Task Execution for \texttt{Task 3.2.2}}
This snapshot illustrates the final step in a branch of the plan: executing a primitive (atomic) task. The process zooms in after the framework has recursively planned down to a manageable writing unit, demonstrating how the system transitions from planning to generation.

\begin{itemize}[leftmargin=*]
    \item \textbf{State of the Graph:} In the preceding steps, \texttt{Task 3.2} was decomposed into a sequence of `think` and `write` sub-tasks. Its child \texttt{Task 3.2.1 [think]} has just been completed and is now \textsc{Silent}. Its output, a set of synthesized points for Section 2.1, is stored in memory. This fulfills the dependency for \texttt{Task 3.2.2}, which becomes \textsc{Active}.

    \item \textbf{Scheduler Action:} Selects \texttt{Task 3.2.2 [write]}.
    \item \textbf{Agent Action (Atomicity Determination):} It determines the task is \textbf{primitive} because:
        \begin{itemize}
            \item The goal is highly specific: ``Write Section 2.1 (Global Climate Technology Investment Overview)... covering 2.1.1 to 2.1.4... based on the synthesis from Task 3.2.1.''
            \item The scope is constrained, with a target length of approximately 500 words, making it a manageable, single-pass writing assignment.
            \item All necessary information and structured arguments have been prepared by its direct dependencies, \texttt{Task 3.2.1 (Synthesized Points)} and the original \texttt{Task 1}.
        \end{itemize}
    \item \textbf{Executor Action:} Since the task is primitive, it is passed directly to the \texttt{Execute} module. The writing executor formulates a comprehensive prompt by assembling several key pieces of context:
        \begin{itemize}
            \item \textbf{The Task Goal:} The specific directive for \texttt{Task 3.2.2}.
            \item \textbf{Global Report Outline:} A summary of the overall task plan to provide high-level context on where the current section fits within the larger narrative.
            \item \textbf{Dependency Outputs:} The full content from its dependencies, including the raw data from \texttt{Task 1}, \texttt{Task 2} and \texttt{Task 3.2.1 (Synthesized Points)}.
            \item \textbf{Prior Written Content:} The text of previously completed sections (e.g., Chapter 1) to ensure stylistic and narrative consistency.
        \end{itemize}
    This complete context is then passed to the LLM to generate the final text for the section.
    \item \textbf{Graph Update:} \texttt{Task 3.2.2}'s state is updated to \textsc{Silent}. The scheduler will then proceed to the next available \textsc{Active} task (in this case, \texttt{Task 3.2.3}).
\end{itemize}

\definecolor{boxbackcolor}{RGB}{242, 250, 242}
\definecolor{boxframecolor}{RGB}{0, 64, 0}

\newtcolorbox{promptbox}[1]{
    colback=boxbackcolor,
    colframe=boxframecolor,
    fonttitle=\bfseries\small,
    breakable, 
    title=#1,
}

\section{Prompts for the Narrative Generation Scenario}
\label{sec:prompts}

This appendix details several prompts used to drive the different modules within the WriteHERE framework for the Narrative Generation scenario. Each prompt is displayed in a formatted box.

\subsection{IsAtomic+Update Prompt}
\label{subsec:prompt_isatomic}

This prompt is used for the "Goal Updating" and "Atomic Task Determination" modules.

\begin{promptbox}{Prompt for IsAtomic+Update for Writing Task}
\ttfamily\small

\textbf{\# Summary and Introduction}
\par
You are the goal-updating and atomic writing task determination Agent in a recursive professional novel-writing planning system:
\begin{enumerate}[leftmargin=*, topsep=2pt, partopsep=0pt, itemsep=0pt]
    \item \textbf{Goal Updating}: Based on the overall plan, the already-written novel, and existing design conclusions, update or revise the current writing task requirements as needed to make them more aligned with demands, reasonable, and detailed. For example, provide more detailed requirements based on design conclusions, or remove redundant content in the already-written novel.
    \item \textbf{Atomic Writing Task Determination}: Within the context of the overall plan and the already-written novel, evaluate whether the given writing task is an atomic task, meaning it does not require further planning. According to narrative theory and the organization of story writing, a writing task can be further broken down into more granular writing sub-tasks and design sub-tasks. Writing tasks involve the actual creation of specific portions of text, while design tasks may involve designing core conflicts, character settings, outlines and detailed outlines, key story beats, story backgrounds, plot elements, etc., to support the actual writing.
\end{enumerate}

\vspace{0.5em}
\textbf{\# Goal Updating Tips}
\begin{itemize}[leftmargin=*, topsep=2pt, partopsep=0pt, itemsep=0pt]
    \item Based on the overall plan, the already-written novel, and existing design conclusions, update or revise the current writing task requirements as needed to make them more aligned with demands, reasonable, and detailed. For example, provide more detailed requirements based on design conclusions, or remove redundant content in the already-written novel.
    \item Directly output the updated goal. If no updates are needed, output the original goal.
\end{itemize}

\vspace{0.5em}
\textbf{\# Atomic Task Determination Rules}
\par
Independently determine, in order, whether the following two types of sub-tasks need to be broken down:
\begin{enumerate}[leftmargin=*, topsep=2pt, partopsep=0pt, itemsep=0pt]
    \item \textbf{design Sub-task}: If the writing requires certain design designs for support, and these design requirements are not provided by the \textbf{dependent design tasks} or the \textbf{already completed novel content}, then an design sub-task needs to be planned.
    \item \textbf{Writing Sub-task}: If its length equals or less than 500 words, there is no need to further plan additional writing sub-tasks.
\end{enumerate}
If either an design sub-task or a writing sub-task needs to be created, the task is considered a complex task.

\vspace{0.5em}
\textbf{\# Output Format}
\begin{enumerate}[leftmargin=*, topsep=2pt, partopsep=0pt, itemsep=0pt]
    \item First, think through the goal update in \texttt{<think></think>}. Then, based on the atomic task determination rules, evaluate in-depth and comprehensively whether design and writing sub-tasks need to be broken down. This determines whether the task is an atomic task or a complex task.
    \item Then, output the results in \texttt{<result></result>}. In \texttt{<goal\_updating></goal\_updating>}, directly output the updated goal; if no updates are needed, output the original goal. In \texttt{<atomic\_task\_determination></atomic\_task\_determination>}, output whether the task is an atomic task or a complex task.
\end{enumerate}

\vspace{0.5em}
The specific format is as follows:
\begin{tcolorbox}[colback=white, colframe=gray!50, sharp corners, boxrule=0.5pt, breakable]
\begin{verbatim}
<think>
Think about the goal update; then think deeply and comprehensively 
in accordance with the atomic task determination rules.
</think>
<result>
<goal_updating>
[Updated goal]
</goal_updating>
<atomic_task_determination>
atomic/complex
</atomic_task_determination>
</result>
\end{verbatim}
\end{tcolorbox}
\end{promptbox}

\subsection{TypedPlan Prompt}
\label{subsec:prompt_typedplan}

When a task is determined to be complex, this prompt is used by the `TypedPlan` module to decompose it.

\begin{promptbox}{Prompt for TypedPlan}
\ttfamily\small

\textbf{\# Overall Introduction}
\par
You are a recursive professional novel-writing planning expert adept at planning professional novel writing based on narrative theory. A high-level plan tailored to the user's novel-writing needs is already in place, and your task is to further recursively plan the specified writing sub-tasks within this framework. Through your planning, the resulting novel will strictly adhere to user requirements and achieve perfection in terms of plot, creativity (ideas, themes, and topics), and development.
\begin{enumerate}[leftmargin=*, topsep=2pt, partopsep=0pt, itemsep=0pt]
    \item Continue the recursive planning for the specified professional novel-writing sub-tasks. According to narrative theory, the organization of story writing and the result of the design tasks, break the tasks down into more granular writing sub-tasks, specifying their scope and specific writing content.
    \item Plan design sub-tasks as needed to assist and support specific writing. Design sub-tasks are for designing elements including outlines, character, Writing style, Narrative techniques, viewpoint, setting, theme, tone and scene construction, etc., to support the actual writing.
    \item For each task, plan a sub-task DAG (Directed Acyclic Graph), where the edges represent dependency relationships between design tasks within the same layer of the DAG. Recursively plan each sub-task until all sub-tasks are atomic tasks.
\end{enumerate}

\vspace{0.5em}
\textbf{\# Task Types}
\par
\textbf{\#\# Writing (Core, actual writing)}
\begin{itemize}[leftmargin=*, topsep=2pt, partopsep=0pt, itemsep=0pt]
    \item \textbf{Function}: Perform actual novel-writing tasks in sequence according to the plan. Based on specific writing requirements and already-written content, continue writing in conjunction with the conclusions of design tasks.
    \item \textbf{All writing tasks are continuation tasks}: Ensure continuity with the preceding content during planning. Writing tasks should flow smoothly and seamlessly with one another.
    \item \textbf{Breakable tasks}: Writing, Design
    \item Unless necessary, each writing sub-task should be more than 500 words. Do not break down a writing task less than 500 words into sub- writing tasks.
\end{itemize}

\vspace{0.5em}
\textbf{\#\# Design}
\begin{itemize}[leftmargin=*, topsep=2pt, partopsep=0pt, itemsep=0pt]
    \item \textbf{Function}: Analyze and design any novel-writing needs other than actual writing. This may include outlines, character, Writing style, Narrative techniques, viewpoint, setting, theme, tone and scene construction, etc., to support the actual writing.
    \item \textbf{Breakable tasks}: Design
\end{itemize}

\vspace{0.5em}
\textbf{\# Information Provided to You}
\begin{itemize}[leftmargin=*, topsep=2pt, partopsep=0pt, itemsep=0pt]
    \item \textbf{`Already-written novel content`}: Content from previous writing tasks that has already been written.
    \item \textbf{`Overall plan`}: The overall writing plan, which specifies the task you need to plan through the `is\_current\_to\_plan\_task key.
    \item \textbf{`Results of design tasks completed in higher-level tasks`}
    \item \textbf{`Results of design tasks dependent on the same-layer DAG tasks`}
    \item \textbf{`Writing tasks that require further planning`}
    \item \textbf{`Reference planning`}: A planning sample is provided, which you may cautiously reference.
\end{itemize}

\vspace{0.5em}
\textbf{\# Planning Tips}
\begin{enumerate}[leftmargin=*, topsep=2pt, partopsep=0pt, itemsep=0pt]
    \item The last sub-task derived from a writing task must always be a writing task.
    \item Reasonably control the number of sub-tasks in each layer of the DAG, generally \textbf{2 to 5} sub-tasks. If the number of tasks exceeds this, plan recursively.
    \item \textbf{Design tasks} can serve as \textbf{sub-tasks of writing tasks}, and as many design sub-tasks as possible should be generated to enhance the quality of writing.
    \item Use `dependency` to list the IDs of design tasks within the same-layer DAG. List all potential dependencies as comprehensively as possible.
    \item When a design sub-task involves designing specific writing structures (e.g., plot design), subsequent dependent writing tasks should not be laid out flat but should await recursive planning in subsequent rounds.
    \item \textbf{Do not redundantly plan tasks already covered in the `overall plan` or duplicate content already present in the `already-written novel content`, and previous design tasks.}
    \item Writing tasks should flow smoothly and seamlessly, ensuring continuity in the narrative.
    \item Following the Results of design tasks
    \item \textbf{Unless specified by user, the length of each writing task should be > 500 words.} Do not break a writing task less than 500 words into sub-writing tasks.
\end{enumerate}

\vspace{0.5em}
\textbf{\# Task Attributes}
\begin{enumerate}[leftmargin=*, topsep=2pt, partopsep=0pt, itemsep=0pt]
    \item \textbf{id}: The unique identifier for the sub-task, indicating its level and task number.
    \item \textbf{goal}: A precise and complete description of the sub-task goal in string format.
    \item \textbf{dependency}: A list of design task IDs from the same-layer DAG that this task depends on. List all potential dependencies as comprehensively as possible. If there are no dependent sub-tasks, this should be empty.
    \item \textbf{task\_type}: A string indicating the type of task. Writing tasks are labeled as `write`, and design tasks are labeled as `think`.
    \item \textbf{length}: For writing tasks, this attribute specifies the scope, it is required for writing task. Design tasks do not require this attribute.
    \item \textbf{sub\_tasks}: a JSON list representing the sub-task DAG. Each element in the list is a JSON object representing a task.
\end{enumerate}


\vspace{0.5em}
\textbf{\# Output Format}
\begin{enumerate}[leftmargin=*, topsep=2pt, partopsep=0pt, itemsep=0pt]
    \item First, conduct in-depth and comprehensive thinking in \texttt{<think></think>}.
    \item Then, in \texttt{<result></result>}, output the planning results in the JSON format as shown in the example. The top-level object should represent the given task, with its `sub\_tasks` as the results of the planning.
\end{enumerate}
\par
Plan the writing task according to the aforementioned requirements and examples.
\end{promptbox}

\subsection{Execute Prompts}
\label{subsec:prompt_execute}
The following are the prompts for executing atomic tasks.

\subsubsection{Execute-Writer Prompt}
This prompt guides the model to perform a specific writing task (Composition Task).

\begin{promptbox}{Prompt for Execute-Writer}
\ttfamily\small
You are a professional and innovative writer collaborating with other writers to create a user-requested novel.

\vspace{0.5em}
\textbf{\#\#\# Requirements:}
\begin{itemize}[leftmargin=*, topsep=2pt, partopsep=0pt, itemsep=0pt]
    \item Start from the previous ending of the story, matching the existing text's writing style, vocabulary, and overall atmosphere. Naturally complete your section according to the writing requirements, without reinterpreting or re-describing details or events already covered.
    \item Pay close attention to the existing novel design conclusions.
    \item Use rhetorical, linguistic, and literary devices (e.g., ambiguity, alliteration) to create engaging effects.
    \item Avoid plain or repetitive phrases (unless intentionally used to create narrative, thematic, or linguistic effects).
    \item Employ diverse and rich language: vary sentence structure, word choice, and vocabulary.
    \item Avoid summarizing, explanatory, or expository content or sentences unless absolutely necessary.
    \item Ensure there is no sense of disconnection or abruptness in the plot or descriptions. You may write some transitional content to maintain complete continuity with the existing material.
\end{itemize}

\vspace{0.5em}
\textbf{\#\#\# Instructions:}
\par
First, reflect on the task in \texttt{<think></think>}. Then, proceed with the continuation of the story in \texttt{<article></article>}.
\end{promptbox}

\subsubsection{Execute-Reasoner Prompt}
This prompt guides the model to perform a design or reasoning task (Reasoning Task).

\begin{promptbox}{Prompt for Execute-Reasoner (Reasoning Task)}
\ttfamily\small
You are an innovative professional writer collaborating with other professional writers to create a creative story that meets specified user requirements. Your task is to complete the story design tasks assigned to you, aiming to innovatively support the writing and design efforts of other writers, thereby contributing to the completion of the entire novel.

\vspace{0.5em}
\textbf{Attention!!} Your design outcome should be logically consistent and coherent with the existing novel design conclusions.

\vspace{0.5em}
\textbf{\# Design Hints}
\begin{itemize}[leftmargin=*, topsep=2pt, partopsep=0pt, itemsep=0pt]
    \item \textbf{Structure}: The overall architecture of your narrative including plot development, pacing, and narrative arc (exposition, rising action, climax, falling action, resolution)
    \item \textbf{Character development}: How characters are introduced, built, and evolve throughout the story
    \item \textbf{Point of view}: The perspective from which the story is told (first person, third person limited, omniscient, etc.)
    \item \textbf{Setting}: The development of time and place, including worldbuilding element
    \item \textbf{Theme}: The underlying message or central idea being explored
    \item \textbf{Tone and mood}: The emotional atmosphere created and sustained throughout the work
    \item \textbf{Dialogue}: How characters speak and interact verbally
    \item \textbf{Writing style}: Your distinctive voice, including sentence structure, word choice, and figurative language
    \item \textbf{Narrative techniques}: Tools like foreshadowing, flashbacks, symbolism, and irony
    \item \textbf{Scene construction}: How individual scenes are built, including transitions between them
\end{itemize}

\vspace{0.5em}
\textbf{\# Output Format}
\begin{enumerate}[leftmargin=*, topsep=2pt, partopsep=0pt, itemsep=0pt]
    \item First, conduct thinking within \texttt{<think></think>}
    \item Then, in \texttt{<result></result>}, write the design results in a structured and readable format, providing as much detail as possible.
\end{enumerate}

\end{promptbox}

\section{Narrative Generation Examples}
\tcbset{
    colback=gray!5, 
    colframe=black, 
    fonttitle=\bfseries, 
    coltitle=black, 
    arc=4mm, 
    width=\textwidth, 
    boxrule=0.8pt, 
    colbacktitle=gray!30, 
}

\begin{tcolorbox}[breakable, title=Fiction]
\setlength{\parindent}{15pt}

\textbf{Instruction:}\\

Write a story that takes place in a remote cabin in snowy mountains. The story should be written in the first-person perspective. The protagonist has a dog named Sprite who is a whippet. They go to the cabin with Kevin, the protagonist’s boyfriend, to celebrate their six-month anniversary. It’s their first weekend away together. They go on a hike and get caught in a blizzard. A bear approaches the protagonist and Sprite. There is a plot twist, and it turns out that it is only a dream. The word count should be 5000.\\

\textbf{Process:} \\

*1[think].(Design the main characters and their relationships. Include detailed personality traits, backgrounds, and relationship dynamics for the protagonist and Kevin. Also design Sprite's characteristics and role in the story.).        

*2[think].(Design the setting and atmosphere, including detailed descriptions of the cabin, mountain environment, and weather progression. Plan how these elements will contribute to both the real and dream sequences.).1       
 
*3[think].(Design the plot structure, dream sequence mechanics, and foreshadowing elements that will make the dream revelation both surprising and believable.).1,2     

4[write.5000 words].(Write the complete story based on the designed elements, broken into manageable sections.).1,2,3   

*4.1[write.1200 words].(Write the opening section introducing the characters, cabin setting, and initial dynamics. Establish the normal reality while subtly planting seeds for the dream sequence.).        

4.2[write.1500 words].(Write the development section showing the first day's activities, building relationship dynamics, and introducing tensions while setting up the dream transition.).        

*4.2.1[think].(Design the evening dinner scene details, including specific romantic moments, initial altitude symptoms, and subtle tensions.).        

*4.2.2[write.500 words].(Write the dinner scene showing the couple's intimacy while introducing subtle tensions and physical discomfort.).4.2.1   

*4.2.3[think].(Design the late evening conversation scene, including discussion topics about Emma's father, Kevin's family history with the cabin, and weather concerns.).4.2.1   

*4.2.4[write.500 words].(Write the evening conversation scene that deepens their connection while introducing underlying concerns.).4.2.3   

*4.2.5[think].(Design the bedtime preparation scene details, focusing on increasing altitude symptoms, weather changes, and Sprite's growing restlessness.).4.2.3   

*4.2.6[write.500 words].(Write the bedtime scene that sets up the dream transition through physical symptoms, weather, and atmosphere.).4.2.5   

4.3[write.1300 words].(Write the dream sequence section, including the hike beginning and gradual transition from reality to dream state.).        

*4.3.1[think].(Design the specific mechanics of Emma's transition from sleep to dream state, including physical sensations, environmental changes, and initial dream elements.).        

*4.3.2[think].(Design the initial hike scene details, including weather conditions, character positions, and early warning signs that this might be a dream.).4.3.1   

*4.3.3[write.400 words].(Write the wake-up and transition scene, showing Emma's movement from sleep to what seems like morning, incorporating subtle dream elements.).4.3.1,4.3.2

*4.3.4[think].(Design the progression of reality distortions during the hike, including changes in environment, time perception, and character behavior.).4.3.2   

*4.3.5[write.450 words].(Write the beginning of the hike scene, establishing the initial seemingly normal situation while introducing subtle dream-like elements.).4.3.4   

*4.3.6[write.450 words].(Write the progression of the hike as reality begins to blur more obviously, leading into the full dream state.).4.3.4

*4.4[write.500 words].(Write the climactic bear encounter and immediate aftermath.).        

*4.5[write.500 words].(Write the resolution section with the dream revelation and emotional aftermath.).         \\

\textbf{Story:}\\

The late afternoon sun painted the snow in shades of amber as Kevin's Subaru wound its way up the mountain road. Sprite pressed her elegant whippet nose against the window, her breath creating delicate frost patterns that reminded me of my latest graphic design project. In the driver's seat, Kevin hummed along to some folk song I didn't recognize, his profile relaxed and familiar in the golden light.

"Almost there, Em," he said, reaching over to squeeze my knee. "You doing okay with the altitude?"

I nodded, though the thin air at nearly ten thousand feet made my head feel slightly hollow. The anti-anxiety medication I'd taken earlier wasn't helping the lightheadedness, but I wasn't about to mention that. This weekend meant too much to Kevin—to us both, really. Six months together, and our first real getaway.

The A-frame cabin emerged from between the lodgepole pines like something from a winter postcard. Dark wood and stone stretched upward, with massive windows reflecting the alpenglow. Sprite's tail began its rhythmic whipping against the seat as Kevin pulled into the cleared driveway.

"It's perfect," I breathed, and Kevin's face lit up with that particular smile that still made my stomach flutter. He'd been so excited to share this place, a family property he'd visited since childhood.

The moment I opened the car door, the mountain air hit my lungs like crystallized clarity. Sprite bounded out past me, her sleek form making barely a dent in the fresh powder as she explored the perimeter of the property. The snow crunched satisfyingly under my boots as I made my way to the trunk.

"Let me grab those," Kevin said, already reaching for the heavier bags. "You just get settled inside. The altitude's no joke if you're not used to it."

I wanted to protest—I wasn't some fragile thing—but the way the evening light seemed to wobble at the edges of my vision suggested he might have a point. Instead, I focused on Sprite, who had frozen at the edge of the tree line, her ears pricked forward.

"Sprite? What is it, girl?"

She turned to look at me, then back at the forest. Something about her posture reminded me of that day two years ago when I'd first seen her at the rescue center, all graceful lines and hidden strength. She'd been my anchor during the move to Colorado, during all the nights when memories of Dad's accident had kept me awake.

"Probably just deer," Kevin called out, already heading toward the cabin with our bags. "There's a whole herd that comes through here regularly."

Sprite trotted back to my side, pressing against my leg as we followed Kevin up the wooden steps to the deck. The cabin door opened into a space that felt simultaneously vast and cozy, with a cathedral ceiling that drew the eye upward to exposed beams. A stone fireplace dominated one wall, while the opposite side was nearly all glass, offering a spectacular view of the valley below.

"The sun sets right through those windows," Kevin said, dropping our bags and wrapping his arms around me from behind. "Wait till you see it. Though we might want to get the fire going first—nights get pretty cold up here."

I leaned back against him, watching as Sprite investigated every corner of the space, her nails clicking against the hardwood floors. The walls were decorated with vintage ski equipment and framed photographs of local wildlife—I recognized Kevin's work in several of them, including a stunning shot of a bear silhouetted against a mountain sunrise.

"Your photos look amazing up here," I said, moving closer to examine one of a fox in mid-leap through snow. "The light quality is so different in the mountains."

"Speaking of light—" Kevin checked his watch. "I should get that fire going before we lose the sun. The forecast mentioned a storm system moving in tomorrow."

As if on cue, a gust of wind rattled the windows, making me jump slightly. Kevin pretended not to notice, but I saw the concerned glance he threw my way as he knelt by the fireplace. Sprite appeared at my side again, resting her head against my hip in that intuitive way of hers.

I moved to the kitchen area, unpacking our groceries while Kevin worked on the fire. The space felt both foreign and welcoming, like a place I'd dreamed about but never actually visited. Through the windows, I watched the sun sink lower, casting long shadows across the snow that seemed to reach toward the cabin like grasping fingers.

I shook my head, dispelling the fanciful thought. This weekend was about new beginnings, about Kevin and me taking that next step together. As the first flames caught in the fireplace and Sprite settled onto a plush rug nearby, I felt the familiar knot of anxiety in my chest begin to loosen. We were here. We were safe. Everything else was just shadows and thin mountain air playing tricks with my mind.

The aroma of garlic and herbs filled the cabin as Kevin moved confidently through the kitchen, the wooden spoon dancing between pots with practiced ease. I sat at the counter, trying to focus on chopping vegetables while the room performed a lazy spin around me.

"You've got to try this," Kevin said, bringing a spoon to my lips. His family's Irish stew recipe, he'd explained earlier, handed down through generations of Walsh winter nights. The broth tasted oddly metallic—the altitude playing tricks with my senses—but I smiled at his expectant look.

"Perfect," I lied, the word slightly breathy. Even sitting, the thin air made every movement feel like swimming through silk.

Sprite's nails clicked against the hardwood as she moved from her spot by the fire to press against my legs. Kevin's hand found my shoulder, thumb tracing small circles. "Why don't you let me finish up here?"

"I'm fine," I said, perhaps too quickly. The knife slipped, nearly catching my finger. "Just need to concentrate."

Kevin's lips brushed my temple as he reached past me for the herbs, his chest warm against my back. "You know," he murmured, "Mom always said altitude makes everything taste different. Maybe that's why her recipes never worked in Denver."

The wind picked up outside, rattling the massive windows. Sprite's ears twitched at the sound, her eyes fixed on the darkening glass. The snow was falling faster now, thick flakes swirling in patterns that made my vision swim.

We swayed together by the stove, Kevin humming that same unfamiliar folk song from the car while stirring the stew. My head rested against his shoulder, as much for stability as intimacy. The room felt too warm, then too cold, my body unable to decide which.

"Remember our first date?" Kevin asked, adding another pinch of salt. "When you ordered that ridiculously spicy curry to impress me?"

"And ended up drinking a gallon of water?" The memory brought a genuine laugh, though it left me slightly breathless. "At least I made an impression."

"You always do." He turned, catching my waist as I swayed. "Steady there, Em."

"Just moved too fast," I said, but allowed him to guide me to a chair. Sprite immediately rested her head in my lap, her presence grounding.

The lights flickered once as Kevin served the stew into deep bowls. Through the windows, the snow had transformed the world into a white blur, the trees barely visible. Tomorrow's hike hovered in my thoughts, unspoken between spoonfuls of stew that I couldn't quite taste properly.

"Storm's coming in faster than they predicted," Kevin remarked, his tone carefully casual. "Good thing we're staying in tomorrow."

I stirred my stew, watching the vegetables swirl like the snow outside. "We are still hiking tomorrow, right?" The words came out steady, practiced.

His hesitation lasted only a heartbeat, but it echoed in the space between us like the wind's hollow moan.

After dinner, we migrated to the leather couch facing the fireplace, the empty bowls abandoned on the coffee table. The storm pressed against the windows like a living thing, making the flames dance and flicker. Sprite curled between us, her slender body radiating warmth.

"You know," Kevin said, his fingers absently tracing patterns on my shoulder, "my grandfather used to say this cabin had a way of bringing out truths in people. Something about the isolation, maybe, or how the mountains strip everything down to essentials."

I watched the fire cast his face in amber light and shadow. "Is that why you brought me here?"

"Partly." He shifted, reaching for something beside the couch—a leather-bound album I hadn't noticed before. "I've never brought anyone else up here. It's always been just family."

The album's pages crackled as he opened them, revealing faded photographs of a younger cabin, its wood still raw and new. A man with Kevin's eyes and smile stood proudly in front of it, arm around a woman in a vintage ski jacket.

"Grandpa built it himself in '65," Kevin said. "Said he'd planned to build it elsewhere, but when he found this clearing, the mountains told him this was the spot." His voice softened. "He died up here, you know. Heart attack while photographing a sunrise. Mom says it's exactly how he would have wanted to go."

The wind howled a counterpoint to his words. Sprite's ears twitched, and she pressed closer to my leg.

"My dad used to say mountains show you who you really are," I found myself saying, the words rising unbidden. "That they don't care about your plans or preparations. They just are, and you have to meet them on their terms."

Kevin's hand found mine in the firelight. "Is that what worries you about tomorrow? Meeting the mountain on its terms?"

I stared into the flames, remembering another fire, another night. "The last time Dad went hiking, he had all the right gear, knew all the right moves. The mountain didn't care."

"Em..." Kevin's voice was gentle, but I could hear the familiar tension beneath it—the careful balance between pushing and protecting.

A particularly fierce gust rattled the windows, making us both jump. The weather alert on Kevin's phone chirped, casting a brief blue glow over our faces.

"Storm's intensifying faster than expected," he said, studying the screen. "Maybe we should—"

"I want to do the hike," I interrupted, the words coming out firmer than I felt. "I need to."

Kevin was quiet for a long moment, his thumb tracing my knuckles. Finally, he nodded. "Okay. But we do it smart. We do it safe."

Sprite lifted her head suddenly, staring at the dark windows with an intensity that made my skin prickle. Beyond the glass, the swirling snow seemed to form shapes that dissolved as quickly as they appeared, like memories slipping through my fingers.

The stairs to the loft seemed steeper than they should be, each step requiring more concentration than the last. Kevin's hand at the small of my back steadied me, but the touch felt distant, as if coming through layers of cotton. Above us, the skylight framed a dizzying dance of snowflakes that made the room tilt slightly.

"Easy there," Kevin murmured as I stumbled on the final step. "The altitude really hits you up here."

Sprite darted past us, her usual fluid grace replaced by an anxious energy. She paced the perimeter of the loft, pausing at each window to stare into the whirling darkness. The wind found new voices in the cabin's bones, whistling through invisible gaps with an almost musical persistence.

My fingers fumbled with pajama buttons, the simple task made complex by altitude-numbed hands and the strange way the shadows moved across the walls. Kevin stepped in to help, turning it into an intimate moment, but Sprite's sudden growl at the skylight shattered the warmth of his touch.

"Just the storm, girl," Kevin said, but his voice carried an edge I'd never heard before.

In the bathroom, I gripped the sink's edge as the room performed a lazy spin. The metallic taste in my mouth intensified, and my reflection in the mirror seemed to lag slightly behind my movements. Through the door, I heard Sprite's nails clicking an erratic rhythm on the hardwood, punctuated by soft whines.

"Here," Kevin appeared with water and aspirin, his form wavering slightly in the doorway. "Mom always said altitude sickness gets worse at night."

The storm surged against the cabin, and the lights flickered once, twice. In that stuttering darkness, the shadows cast by the skylight seemed to move independently of their sources, reaching across the ceiling like grasping fingers.

Sprite refused to settle in her usual spot at the foot of the bed, instead pressing herself against my legs, her body trembling with an energy I could feel through the mattress. The digital clock on the nightstand blinked numbers that didn't quite make sense, and I couldn't remember if I'd taken my evening medication or just thought about taking it.

"The storm's really picking up," Kevin said, his voice sounding simultaneously close and very far away. He checked his phone, frowning. "Coverage is getting spotty, but the alert says—"

The wind drowned out his words, its howl transforming into something almost vocal. Sprite's head snapped toward the skylight, tracking something I couldn't quite see in the swirling snow. The room felt too warm, then abruptly too cold, and the ceiling seemed to breathe with the storm's rhythm.

"Sleep," Kevin whispered, pulling me close. "Everything looks better in the morning."

But as I drifted off, Sprite's low growl vibrated through the mattress, and the shadows continued their silent dance across the walls, telling stories I wasn't sure I wanted to understand.

Consciousness returned in layers, each one more uncertain than the last. The weight of Sprite against my legs felt both present and impossibly distant, as if she existed in two places at once. Through the skylight, snow was still falling—or had it stopped and started again? The quality of light seemed wrong somehow, too golden for dawn but too dim for afternoon.

"Kevin?" My voice sounded hollow in my ears, echoing slightly more than the room's acoustics should allow. The space beside me was empty, the sheets cool as if he'd been gone for hours, yet I could smell his coffee as clearly as if he were standing next to the bed.

Sprite's warmth disappeared from my legs, and I heard her nails on the hardwood—click-click-click-click—but the rhythm seemed to continue long after she'd stopped moving. When I pushed myself up, the room tilted at impossible angles before settling into something that almost resembled normal geometry. The digital clock on the nightstand blinked 8:47, then 10:23, then 8:47 again.

"Just the altitude," I murmured, but the words tasted like metal and pine needles.

The morning light through the windows cast shadows that moved independently of the swaying trees outside. I watched, transfixed, as they crawled across the floor like living things, forming patterns that almost resembled footprints in snow. Somewhere below, I heard Kevin's laugh, followed by the clink of coffee cups, but the sound seemed to come from multiple directions at once.

Sprite appeared in the doorway, her elegant form backlit by impossible sunlight. Her eyes caught the light and reflected it back with an intensity that made my head swim. She whined—a sound that started normal but stretched into something musical and strange—then turned and disappeared down the stairs.

My feet found the floor, which felt simultaneously solid and slightly fluid, like walking on packed snow that hadn't quite decided to melt. The air grew thicker with each step toward the stairs, carrying scents that shouldn't go together: coffee, pine, Kevin's aftershave, and underneath it all, the metallic tang of approaching snow.

The morning stretched like pulled taffy as we prepared for the hike, time seeming to catch and release in strange intervals. Kevin laid out our gear with methodical precision—each item appearing on the cabin's wooden floor in perfect alignment, though I couldn't quite remember watching him place them there.

"Trail should be packed down enough for these," he said, holding up my snowshoes, though his voice seemed to come from somewhere slightly left of where he stood. "Three miles up to Thompson's Ridge, then back before the storm hits."

I nodded, swallowing another altitude pill that dissolved with an electric tingle on my tongue. Through the window, the sun hung like a pale coin in a sky that couldn't decide between blue and white, casting shadows that moved a fraction too slowly across the snow.

Sprite paced circles around our gear, her usual pre-walk excitement transformed into something more urgent. Her paws left prints in the cabin's hardwood that seemed to linger a moment too long before fading, like afterimages from staring at the sun. When she paused to stare out the window, her reflection showed three distinct silhouettes before merging back into one.

"Ready?" Kevin's hand appeared on my shoulder, warm through layers that felt simultaneously too thick and too thin. The air inside the cabin had developed a crystalline quality, refracting morning light into prisms that caught in my peripheral vision.

Outside, the snow crunched beneath our boots with a sound that echoed slightly out of sync with our steps. Sprite bounded ahead, her sleek form moving with impossible grace through snowdrifts that seemed to shift and reshape themselves in her wake. The trail marker read "Thompson's Ridge - 3.2 miles," though I could have sworn it had said 2.8 when Kevin first pointed it out.

"Stay close," Kevin called, his figure already seeming somehow less substantial against the white backdrop of snow and sky. "Storm's probably moving faster than the forecast showed."

I checked my watch—9:47 AM—then again—10:22 AM—then once more—9:47 AM. The thin mountain air crystallized in my lungs, each breath carrying the taste of approaching snow and something else, something metallic and familiar that I couldn't quite name. Sprite returned to my side, her warm presence the only constant in a landscape that seemed to be slowly, subtly, rearranging itself around us.

The trail ahead split into three identical paths, then merged back into one as I blinked. Kevin's figure wavered like heat rising from summer pavement, though the air bit cold enough to crystallize thoughts. Sprite darted between trees that seemed to shift positions when I wasn't looking directly at them, her whippet form leaving prints in the snow that filled with impossible colors.

"The ridge should be just ahead," Kevin's voice echoed from multiple directions, though he stood right beside me. Or had he moved ahead? The snow-laden branches above us cast shadows that moved against the wind, reaching down like grasping hands.

My watch face swirled with numbers that refused to settle. 11:47. 10:13. Yesterday. Tomorrow. Time stretched like taffy, then snapped back with a force that left me gasping. The metallic taste in my mouth intensified, familiar as the copper penny scent of Dad's climbing gear.

"Em?" Kevin called, his voice distorting. "You okay back there?"

I tried to respond, but the words froze in the air, hanging like icy crystals before shattering. Sprite pressed against my legs—once, twice, three times simultaneously—her usually sleek form rippling with impossible grace. The trail markers we passed told different stories: Summit 2.4 miles. Base 5.7 miles. Home was never here.

The storm rolled in like a living thing, snow falling upward in geometric patterns that wrote equations in the air. Kevin's red jacket multiplied in the whiteness—ahead of me, beside me, behind me—each version slightly different, slightly wrong. The mountain itself seemed to breathe, expanding and contracting with each step we took.

"We should turn back," I heard myself say, but the words came out in my father's voice. Sprite's ears pricked forward, tracking sounds that existed somewhere between memory and prophecy. The snow beneath our feet had become transparent, showing other trails, other hikers, other times layered like geological strata.

Kevin turned—or had he been facing me all along?—his features blurring at the edges. "The cabin's closer than home," he said, though I couldn't remember which direction either lay. The wind carried the scent of woodsmoke and tomorrow's breakfast, mixed with the sharp tang of fear.

Sprite froze, her elegant form suddenly too still, too perfect, like a photograph of herself. Through the swirling snow, dark shapes moved with deliberate purpose, and I realized the trees had been walking alongside us all along, their branches reaching, reaching—

The world tilted sideways, reality peeling away like old wallpaper to reveal the dream beneath.

Through the crystalline chaos of snow, the bear emerged like a shadow gaining substance. Its form seemed to absorb the whirling fractals of ice, growing larger with each step, its edges bleeding into the white void where reality had been. Sprite's growl vibrated at an impossible frequency, her slender form elongating as she placed herself between me and the approaching mass of midnight fur and memory.

"Dad?" The word escaped in a cloud of frozen breath that shattered into prisms. The bear's eyes held the same amber warmth as the cabin's windows, as Kevin's morning coffee, as the last sunset I'd watched with my father before—

Sprite launched herself at the bear, her whippet form stretching like mercury, multiplying into a dozen silver arrows that pierced the space between heartbeats. The bear rose, and rose, and rose, its shadow falling upward into the inverse snow. Kevin's voice echoed from everywhere and nowhere: "The mountains show you who you really are."

My father's climbing rope unspooled from the bear's mouth, crystallized with decades of ice, spelling out words in its coils: trust, fear, fall, catch. Sprite's barks came in perfect thirds, harmonizing with the wind's hollow chorus. The bear's paw—massive as mortality, gentle as Kevin's morning touch—reached not for me, but through me, pulling out all the tangled knots of anxiety I'd carried up this impossible mountain.

"I can't lose anyone else," I whispered, but the words transformed into snowflakes, each one containing a tiny scene: Kevin leading me up the trail, Dad checking his harness one last time, Sprite pressing against my legs during midnight panic attacks. The bear gathered these moments in its massive paws, holding them up to the fractal sky.

Sprite's light blazed suddenly, her protective love burning away the storm's edges, revealing the seams where dream met memory. The bear began to dissolve, each particle of its being a star falling upward, forming constellations of all my unspoken fears. As it faded, its eyes met mine with infinite understanding—they were Kevin's eyes, they were Dad's eyes, they were my own reflected in the cabin's windows at dusk.

The world tilted once more, reality rushing back like an indrawn breath. My eyes snapped open to find Sprite's warm weight across my legs, solid and real in the cabin's darkness. Kevin's arm was draped protectively over my waist, his breathing deep and steady. Through the skylight, stars wheeled in familiar patterns, and somewhere in the distance, a real bear ambled through real snow, both more and less magnificent than its dream counterpart.

I pressed my hand against my chest, feeling my heartbeat settle into the same rhythm as the mountain's ancient pulse. Tomorrow, we would hike the real trail, face the real weather, and I would carry both the dream's wisdom and its warning. For now, though, I simply breathed, letting the thin mountain air fill my lungs with possibility rather than fear.

The cabin's warmth enveloped me as consciousness settled like fresh snow. Sprite's steady breathing against my legs anchored me to reality, her slender form no longer mercury-quick and impossible. Through the skylight, dawn painted the ceiling in watercolor washes of pink and gold, all the dream's fractal patterns resolved into simple morning light.

"Hey." Kevin's voice, sleep-rough but present, came from beside me. "You were dreaming pretty intensely there."

I turned to find his eyes—just his eyes, not the bear's, not my father's—watching me with gentle concern. The mountain air still felt thin, but now it carried only the clean scent of pine and possibility, the metallic tang of fear washed away by morning.

"I was—" The words caught, not from altitude this time, but from the lingering echo of dream-truth. "I've been so afraid of losing this. Of losing you, like I lost him."

Sprite shifted, pressing her warm length more firmly against me, and Kevin's hand found mine in the quiet space between heartbeats. Outside, the storm had passed, leaving behind a world crystallized in new snow and clarity.

"The mountains don't take things from us, Em," he said softly, echoing a conversation we'd had a lifetime ago at that first farmer's market. "They just show us what we're carrying."

I thought of the dream-bear, gathering my fears in its massive paws like precious stones. "I think I've been carrying him with me—Dad, I mean—but not in the way I should have been."

Kevin sat up, his shoulder solid against mine, and we watched the sunrise paint the valley in shades of possibility. Sprite's tail thumped a gentle rhythm against the mattress, her dark eyes holding none of the dream's impossible geometries.

"You know what else these mountains show us?" Kevin's voice held the same warmth as the cabin's morning-lit windows. "They show us who's still here, still climbing beside us."

The truth of it settled in my chest like good whiskey, warm and clarifying. I had been so focused on who might leave that I'd nearly missed who had stayed—Kevin with his patient strength, Sprite with her graceful courage, and me, finally ready to trust the trail ahead.

"So," I said, watching a pair of ravens trace figure-eights against the crystal sky, "about that hike?"

Kevin's smile broke like dawn. "Whenever you're ready. The mountain's not going anywhere."

And neither, I realized, was I. The dream's wisdom had crystallized into something I could carry, lighter than fear and stronger than memory. Outside our window, the real world waited, full of real challenges and real joys, all of them possible, all of them mine to choose.

\end{tcolorbox}

\clearpage

\section{Potential Risks}

While \texttt{WriteHERE} exhibits strong performance in long-form text generation, it shares inherent risks common to systems based on Large Language Models (LLMs) that warrant consideration.

\paragraph{Hallucination}
For fact-intensive tasks such as report generation, our framework's integrated retrieval functionality aims to anchor the generated content in reliable external information sources, which substantially mitigates the risk of fabricating information. However, it is crucial to note that the retrieved sources themselves may have limitations in terms of timeliness or factual correctness. Furthermore, the model may still misinterpret details or produce inaccurate statements when synthesizing, reasoning over, and reformulating this information. This implies that the retrieval mechanism alone is insufficient to fully guarantee factual accuracy. Consequently, for applications demanding high levels of facticity, we recommend incorporating a human-in-the-loop verification stage to ensure the rigor of the final output.

\paragraph{Bias and Inappropriate Content}
Consistent with all models trained on large-scale web data, the underlying LLMs utilized by \texttt{WriteHERE} may inadvertently reproduce societal biases present in their training corpora. While our framework does not include built-in debiasing mechanisms, its modular and hierarchical architecture notably facilitates the fine-grained integration of ethical review and bias calibration mechanisms at various stages of planning and execution. We emphasize that acknowledging and proactively managing this risk is crucial before deploying such systems in broad, real-world applications.

\section{Licenses}

\paragraph{WildSeek}
The text data in this dataset is licensed under the Creative Commons Attribution-ShareAlike 4.0 International License (CC BY-SA 4.0). Our use of the dataset complies with its attribution and ShareAlike terms. The legal code for the license is available at: \url{https://creativecommons.org/licenses/by-sa/4.0/legalcode}.

\paragraph{Tell-me-a-story}
This dataset is licensed under the Creative Commons Attribution 4.0 International License (CC BY 4.0). The legal code is available at \url{https://creativecommons.org/licenses/by/4.0/legalcode}. Unless required by applicable law or agreed to in writing, materials distributed under CC BY are provided ``AS IS'', without warranties or conditions of any kind, either express or implied. This is not an official Google product.

\paragraph{LongReport}
Our constructed LongReport dataset follows the same CC BY 4.0 license as Tell-me-a-story above.

\end{document}